%% file: root.tex
\documentclass[journal]{IEEEtran} 
\usepackage{multirow}
\usepackage{cite}
\usepackage{hyperref}
\usepackage[dvipsnames]{xcolor}
\hypersetup{    
    colorlinks=true, 
    linkcolor=NavyBlue,
    urlcolor=NavyBlue,
    citecolor=Mahogany     
}
\usepackage[normalem]{ulem}
\usepackage{graphicx} 
\usepackage{amsmath} 
\usepackage{amssymb}  
\usepackage{mathrsfs}  

\usepackage{amsthm}  
\usepackage{circledsteps}
\usepackage{adjustbox}
\usepackage[ruled,linesnumbered]{algorithm2e}
\usepackage{algpseudocode}

\include{commands}

\title{\LARGE \bf
Safe multi-agent motion planning under uncertainty for drones using filtered reinforcement learning}

\author{
Sleiman Safaoui$^\dagger$, Abraham P. Vinod$^{\dagger\ast}$, Ankush Chakrabarty, Rien Quirynen, Nobuyuki Yoshikawa,\linebreak 
and Stefano Di Cairano
\thanks{
    $^\dagger$ Equal contributions.
    \newline\indent $^\ast$ Corresponding author. Email: \texttt{abraham.p.vinod@ieee.org}.\newline\indent
    S. Safaoui is with Eric Jonsson School of Engineering \& Computer Science, The University of Texas at Dallas, Richardson, TX, 75080, USA. This work was completed while he was at Mitsubishi Electric Research Laboratories.\newline\indent 
    A. Vinod, A. Chakrabarty, R. Quirynen, and S. Di Cairano are with Mitsubishi Electric Research Laboratories, Cambridge, MA 02139, USA. \newline\indent  
    N. Yoshikawa is with Mitsubishi Electric Corporation, Japan.}%
}

\begin{document}

\maketitle
\thispagestyle{empty}
\pagestyle{empty}

\begin{abstract}
    We consider the problem of safe multi-agent motion planning for drones in uncertain, cluttered workspaces. For this problem, we present a tractable motion planner that builds upon the strengths of reinforcement learning and constrained-control-based trajectory planning. First, we use single-agent reinforcement learning to learn motion plans from data that reach the target but may not be collision-free. Next, we use a convex optimization, chance constraints, and set-based methods for constrained control to ensure safety, despite the uncertainty in the workspace, agent motion, and sensing. The proposed approach can handle state and control constraints on the agents, and enforce collision avoidance among themselves and with static obstacles in the workspace with high probability. The proposed approach yields a safe, real-time implementable, multi-agent motion planner that is simpler to train than methods based solely on learning. Numerical simulations and experiments show the efficacy of the approach.
\end{abstract}


\section{Introduction}\label{sec:intro}
Multi-agent motion planning in cluttered workspaces under stochastic uncertainty arising from both
perception and actuation is a key challenge in designing
reliable autonomous systems. The need for such planners{, especially for quadrotors,} arises in a
variety of application areas including transportation,
logistics, monitoring, and agriculture.
Recently, motion planning using reinforcement learning (RL)
has gained attention, due to its ability to leverage data to tackle generic dynamical systems and complex task specifications~\cite{qie2019joint, lv2019path,  lowe2017multi,Imitationlearning,everett2018motion, semnani2020multi,levine2016end,cheng2019end,zhang2021multi}.
A major challenge for such efforts is the lack of safety
guarantees, since most of the existing RL-based approaches
enforce safety constraints by soft constraints and are subject to training errors.
Additionally, multi-agent {RL} is known to
suffer from non-stationarity and scalability issues, which
may prevent the training to converge~\cite{zhang2021multi}.
We propose \emph{a tractable approach to safe, multi-agent motion planning in stochastic, cluttered workspaces that
combines reinforcement learning and set-based methods for constrained control. Our approach yields a safe, real-time
implementable, multi-agent motion planner that is simple to train and enforces safety with high probability, by means of chance constraints.}
\begin{figure}[t]
    \centering
    \includegraphics[width=0.8\linewidth]{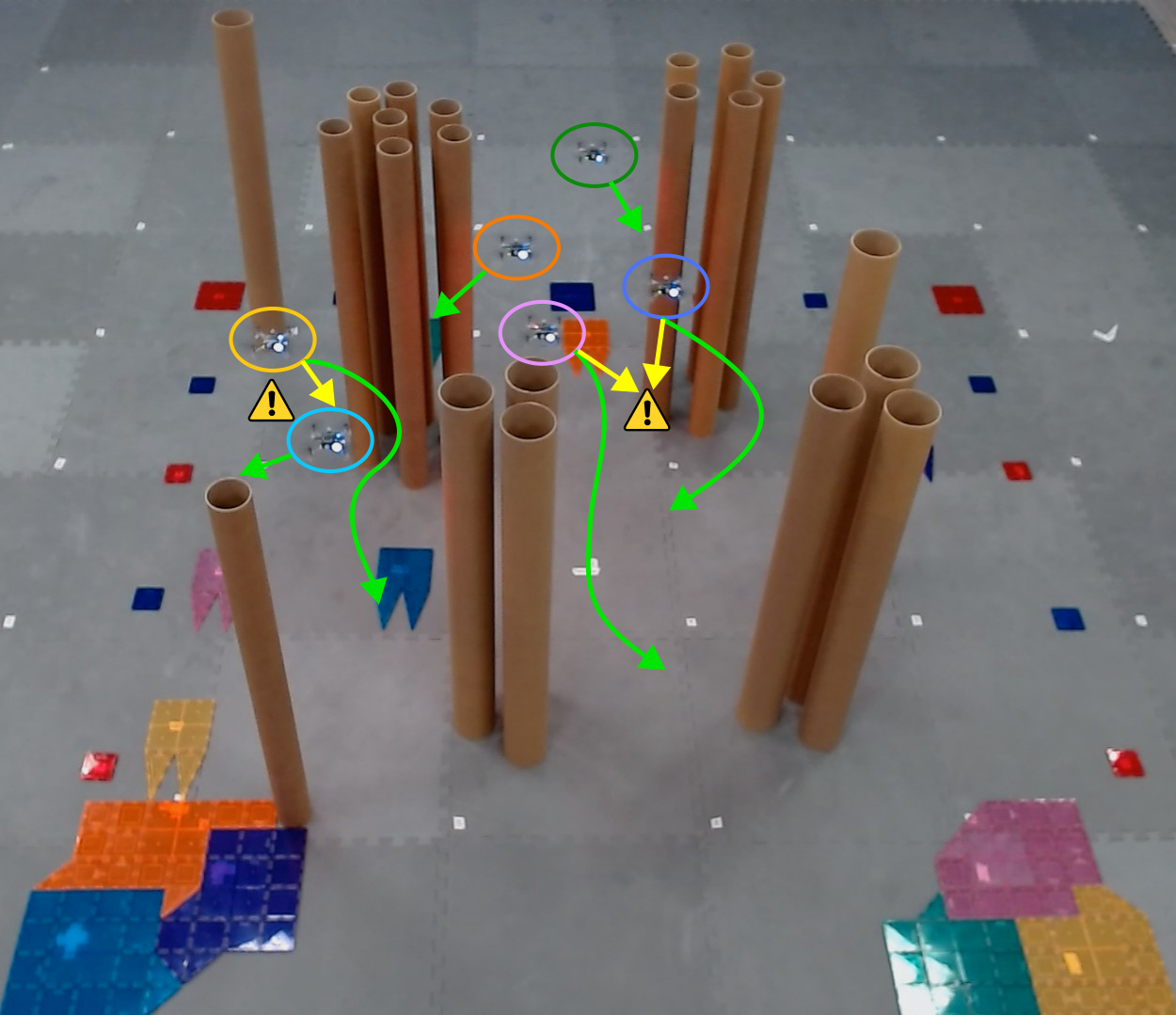}
    \caption{Existing reinforcement learning-based motion
        planners can generate unsafe trajectories (yellow
        arrows), since they treat safety as soft
        constraints, which is undesirable in safety-critical
        applications. We propose a constrained-control-based
        safety filter that renders such motion planners safe
        (long green arrows) by enforcing safety as hard
        constraints. {See~\url{https://youtu.be/QcCSYNwuuo8} for an overview and videos of the experiments.}} 
    \label{fig:exp}
\end{figure}

In the deterministic setting, various approaches have been proposed for multi-agent motion planning such as 
centralized scheduling and coordination~\cite{gravell2021centralized}, 
roadmap and discrete search followed by trajectory refinement~\cite{honig2018trajectory},  
sampling-based rapidly-exploring random trees~\cite{ragaglia2016multi}, 
adaptive roadmaps~\cite{sud2008real}, 
buffered Voronoi cells~\cite{7828016}, 
mixed-integer programming~\cite{radmanesh2016flight}, 
sequential convex programming~\cite{chen2015decoupled, augugliaro2012generation}, 
formal methods and finite transition systems~\cite{verginis2017decentralized}, and
control barrier
functions~\cite{wang2017safety,srinivasan2018control,cheng2019end}.
Recently, RL-based planners have been used in complex environments~\cite{qie2019joint, lv2019path,   lowe2017multi,Imitationlearning,everett2018motion, semnani2020multi,levine2016end,cheng2019end,zhang2021multi}.  
A key advantage of learning-based planners is the ability to leverage past experience in future decision making~\cite{sutton2018reinforcement}. Consequently, such planners can tackle \emph{complex} and \emph{high-dimensional} motion planning tasks, while incorporating prior information about the planning task and accommodating uncertainty~\cite{zhang2021multi}.

Our preliminary work~\cite{our_icra} considered {\emph{deterministic}} safe
multi-agent motion planning, where everything was known exactly. We proposed a two-step approach where a
single-agent RL algorithm provided a \emph{reference} command which was subsequently  \emph{filtered} (or corrected) by a constrained control module. The works closest to~\cite{our_icra} are~\cite{rodionova2020learning,rodionova2021learning,cai2021safe,elsayed2021safe} that also follow a similar two-step process.
However,~\cite{rodionova2020learning,rodionova2021learning}
need labelled data for supervised learning, and
~\cite{cai2021safe,elsayed2021safe} use multi-agent RL.
Multi-agent RL trains multiple agents to collectively
complete the task, and as a consequence, is harder to train than
single-agent RL. Additionally,~\cite{elsayed2021safe} relies on solving a two-player
game in problems with discrete state and action space. 
Our approach utilizes single-agent RL that is simpler to train, can accommodate continuous state and action spaces for
stabilizable linear dynamics, is real-time implementable, and need not be retrained as the number of agents increase. 

In the stochastic setting, the presence of probabilistic constraints makes the motion planning problem more challenging.
Here, we must balance the conservativeness of the motion plans with the risk (the probability of violation of a
desirable property), while ensuring the existence of a solution that satisfies other problem objectives. The authors of~\cite{blackmore2011chance,vinod2018stochastic} propose single-agent motion planners using chance constrained programming, but these approaches may become prohibitively expensive when extended to multi-agent systems.
~\cite{malone2017hybrid} combined artificial potential fields with stochastic reachability theory to generate motion plans for a single-agent in stochastic, cluttered workspaces.
Recently,~\cite{wabersich2018linear,wabersich2021predictive} proposed using a \emph{reference} controller, such as an RL controller trained offline, followed by a constrained-control-based online filtering step to guarantee safety of the control action being applied to the system,
which was applied to autonomous racing~\cite{tearle2021predictive}.
However, to the best of our knowledge, these works do not consider a multi-agent setup.
Another line of research for multi-agent motion planning in stochastic settings uses buffered Voronoi cells~\cite{zhu2022decentralized}, where motion plans are restricted to safety sets computed based on the ``best'' separating hyperplane between two Gaussian distributions, further tightened by a safety buffer. Such an approach couples planning and safety control, yet it does not guarantee recursive feasibility.

We focus on safe multi-agent motion planning in applications where a \emph{centralized} decision maker coordinates the
actions of the agents for safety and performance. Examples of such applications include air traffic control and
coordinated traffic control centers~\cite{Firoozi2022jun}. The  proposed centralized approach may impose additional
communication and computational burden when compared to a decentralized method (for e.g., ~\cite{zhu2022decentralized}).
However, the ability to enforce coordination helps the proposed approach typically generate safer and more efficient
trajectories for the overall system. 

\textbf{Contributions of this work:}  
{
Since RL has been recently of interest to the robotics community, we propose a solution to address the lack of safety in RL-based planners, specifically in multi-agent motion planning settings.
We propose an optimization-based safety filter that, when used in conjunction with RL, provides
a safe, multi-agent motion planner in cluttered
workspaces under stochastic uncertainty. The proposed safety filter uses
convex optimization and set-based control to compute minimum-norm 
corrections to the RL-based motion plan and guarantee probabilistic collective safety of the multi-agent system.
We use single-agent RL to learn from data, while avoiding issues like non-stationarity and scalability that affect multi-agent RL.}
We also describe how to design terminal state constraints for the constrained-control-based safety filter by using reachability to achieve recursive feasibility. Finally, we demonstrate our approach by both numerical simulations and experiments using quadrotors.

We note that while the proposed solution is discussed in the context of RL, our approach is general and can be used with another planner instead of RL.
For instance, the single-agent planner could be sampling-based (e.g. RRT-based planners~\cite{lavalle1998rapidly, karaman2011sampling}), and the advantage of our architecture would be in the dimensionality reduction and reduced effort for collision checking with respect to applying the sampling-based planner to the full multi-agent problem.

\textbf{Relationship with our preliminary work~\cite{our_icra}}: In~\cite{our_icra}, we proposed a safe, multi-agent motion
planner using reinforcement learning and optimization for deterministic dynamics and a known workspace. We generated
continuous-time safety guarantees under the assumption that the safety filter's control input is
constant across the entire horizon of the safety filter.  In this work, we extend
our preliminary work~\cite{our_icra} to stochastic workspace, dynamics, and sensing, and provide probabilistic safety
guarantees for the overall system without relying on the constant control input assumptions. We also explicitly assess
recursive feasibility of the safety filter, and investigate the importance of the RL controller, safety filter, and the
terminal constraints in the proposed approach using extensive hardware and simulation experiments.

\subsection{Notation} \label{sec:notation}
$0_d$ {($0_{n,m}$)} is a vector {(matrix)} of zeros in $\bbr^d$ {($\bbr^{d\times m}$)},
$I_d$ is the $d$-dimensional identity matrix, 
and $\Nint{a}{b}$ is the subset of natural numbers between (and including) $a,b\in \bbn$, $a \leq b$, and
$\Nint{a}{b}=\emptyset$ when $a>b$. 
$\oplus, \ominus$ are the Minkowski sum and Pontyagrin difference, respectively, 
and $\| \cdot \|$ is the 2-norm of a vector. 
The support function of a convex and compact set $ \cvxCompactSet$ is $S_{\cvxCompactSet}(\ell)\triangleq\sup_{x\in \cvxCompactSet} \ell \cdot x$ for any $\ell \in \bbr^d$~\cite{kolmanovsky1998theory}.

$(\Omega, \calf, \bbp)$ is a probability space where $\Omega$ is the sample space, $\calf$ is a $\sigma$-algebra of
subsets of $\Omega$, and $\bbp$ is a probability measure on $\calf$.  We denote random vectors in bold $\rv{x}: \Omega
\rightarrow \bbr^n$ and their mean $\mean{x}\triangleq\bbe[\rv{x}]$, where $\bbe$ is the expectation operator with
respect to $\bbp$.  We use $\estrealize{x}$ to denote a realization of a random vector $\rv{x}$.  We use $\caln(\mu,
\Sigma)$ to denote a Gaussian random vector with mean $\mu$ and covariance $\Sigma$, and refer to $\caln(0_n, I_n)$ as
the standard Gaussian random vector.  For a vector $\boldsymbol{v}(t)$, $\boldsymbol{v}(k|t)$ is the predicted value at
$k\geq t$ based on the information available at time $t$, and we denote
$\boldsymbol{v}(t|t)=\boldsymbol{v}(t)$. We use the same notation when referring to the distribution of
$\boldsymbol{v}(t)$.

The following abbreviations are used throughout the paper: iid (independent and identically distributed), MPC (model predictive control), QP (quadratic program), and PSD (positive semidefinite).

\section{Problem formulation} \label{sec:prob_formulation}
\textbf{Dynamics:} 
Consider $\numAgents\in \bbn$ homogeneous agents with the discrete-time linear dynamics at time $t$,
\begin{align}
    \rv{x}_i(t+1) = A \rv{x}_i(t) + B u_i(t) + \rv{w}_i(t),\label{eq:stoch_dt_time_dyn}
\end{align}
with state $\rv{x}_i\in \bbr^n$, 
input $u_i\in \inputSet \subset \bbr^m$, 
process noise $\rv{w}_i \in \bbr^n$,
state update matrix $A\in \bbr^{n\times n}$, and input matrix $B\in \bbr^{n\times m}$ for each agent $i\in\Nint{1}{N}$. 
The input set $\inputSet$ is a convex and compact polytope,
and the process noise is iid, zero-mean Gaussian $\rv{w}_i
\sim \gauss{0_n, \Sigma_w}$ for some known PSD matrix
$\Sigma_w \in \bbr^{n\times n}$. The position of the agent $i$ at time $t$ is given by,
\begin{align}
    \rv{p}_i(t)= C\rv{x}_i(t),\label{eq:pos}
\end{align}
for some $C\in \bbr^{d\times n}$, $d<m$. For $k\geq t$, the mean state of the agents is predicted according to the nominal dynamics,
\begin{subequations}
\begin{align}
    \mean{x}_i(k+1|t) &= A \mean{x}_i(k|t) + B u_i(k|t)\\
    \mean{p}_i(k|t) &= C\mean{x}_i(k|t).
\end{align} \label{eq:mean_dynamics}%
\end{subequations}%
We assume that the nominal dynamics \eqref{eq:mean_dynamics} are
stabilizable, i.e., there exists a stabilizing gain matrix
$K\in \bbr^{m\times n}$ that ensures that all
eigenvalues of $(A+BK)$ lie in the unit circle.

\textbf{Measurement model:} 
We assume that the initial state of each agent
$\rv{x}_i(0)={x}_i(0)$ is known, i.e.,
deterministic.  However, for $t > 0$, we have access only to
a noisy measurement of the true state $\rv{x}_i(t)$.
Specifically, we assume that the measurements
$\estrealize{y}_i(t)$ are a realization of a random
vector $\rv{y}(t)$,
\begin{align}
    {\rv{y}_i}(t)=\rv{x}_i(t)+\rv{\eta}_i(t),\label{eq:measurement_model}
\end{align}
where $\rv{\eta}_i \sim \gauss{0_n, \Sigma_\eta}$ is a zero-mean Gaussian noise with a known PSD matrix $\Sigma_\eta\in\bbr^{n\times n}$.

\textbf{Agent Representation:} We consider agents with identical convex and compact rigid bodies, denoted by $ \agent \subset \bbr^d$, such that $0_d \in \agent$. The rigid bodies of the agents are rotation-invariant~\cite{lavalle2006planning}. So, we only consider translations.

\begin{rem}
    We can generalize all the presented results to account for heterogeneous agents with heterogeneous linear dynamics, measurement models, and rigid bodies. We consider homogeneity in all of these aspects to simplify the presentation. 
\end{rem}

\textbf{Workspace Representation:} We represent the workspace using a convex and compact polytope $\envPoly \subset \bbr^d$.

\textbf{Obstacle Representation:} 
The workspace has $\numObst$ static obstacles, each with a convex and compact rigid body $\obst{j}\subset \bbr^d$ and $0_d \in \obst{j}$ ($j \in \Nint{1}{\numObst}$). The obstacle shapes are known \emph{a priori}, but their positions are available only via a noisy measurement. Specifically, for each obstacle $j \in \Nint{1}{\numObst}$, the position of a representative point of the obstacle (e.g. the \emph{center}) is denoted by $\rv{c}_j \in \bbr^d$ where $\rv{c}_j$ is an iid Gaussian random vector $\rv{c}_j \sim \caln (\mean{c}_j, \Sigma_{c_j})$ with nominal position $\mean{c}_j$ and covariance matrix $\Sigma_{c_j}\in\bbr^{d\times d}$.

\subsection{Safe multi-agent motion planning under uncertainty} \label{sec:stoch_safety_prob}
Given target positions $q_i\in \bbr^d$, we want to design a motion planner that drives the agents towards their
respective target positions, while ensuring safety of the agents at all times, despite the uncertainty in the dynamics
and the noisy estimates of the agent and obstacle positions. Here, we formalize the required features of safety in the
multi-agent motion planning problem by introducing the notion of \emph{probabilistic collective safety}, inspired by
existing literature~\cite{our_icra,lavalle2006planning}.

\begin{defn}[\textsc{Probabilistic Collective Safety}] \label{def:prob_collective_safety}
    The agents are said to be \emph{probabilistically collectively safe} at time $t$ when all the following criteria are met:
    \begin{enumerate}
        \item \emph{Static obstacle avoidance constraints}: The probability of collision of agent $i\in \Nint{1}{\numAgents}$ with obstacle $j\in \Nint{1}{\numObst}$ is less than a pre-specified risk bound $\alpha_{i,j,t}\in(0,1)$,
        \begin{align}
         \bbp((\rv{p}_i(t) \oplus \agent)\cap (\rv{c}_j(t) \oplus \obst{j}) \neq \emptyset) &\leq \alpha_{i,j,t}. \label{eq:prob_collect_static_obstacle_cc}
        \end{align}
        \item \emph{Inter-agent collision avoidance}: The probability of collision between agents $i,i'\in \Nint{1}{\numAgents},\ i\neq i'$ is less than a pre-specified risk bound $\beta_{i,i',k}\in(0,1)$,
        \begin{align}
            \bbp((\rv{p}_i(t) \oplus \agent)\cap (\rv{p}_{i'}(t) \oplus \agent) \neq \emptyset) \leq \beta_{i,i',t}.
            \label{eq:prob_collect_inter_agent_cc}
        \end{align}
        \item \emph{Keep-in constraints}: The probability of agent $i\in \Nint{1}{\numAgents}$ exiting the keep-in set $\envPoly$ is less than a pre-specified risk bound $\kappa_{i,k}\in(0,1)$,
        \begin{align}
        \bbp(\rv{p}_i(t) \oplus \agent \nsubseteq \envPoly) \leq \kappa_{i,k}.    
        \label{eq:prob_collect_keepin_cc}
        \end{align}
    \end{enumerate}
\end{defn}
Next, we formulate the problem tackled in this paper.
\begin{prob}[\textsc{Safe Multi-Agent Planning}] \label{prob:safe_stoch_multi_agent_planning}
     Given user-specified risk bounds $\alpha_{i, j, t},\
     \beta_{i, i', t},\ \kappa_{i, t}$, for every $i\in
     \Nint{1}{\numAgents},\ i'\in
     \Nint{1}{i-1}, \  j \in
     \Nint{1}{\numObst}$ and $t\in\bbn$, design a multi-agent motion
     planner that navigates the agents with dynamics
     \eqref{eq:stoch_dt_time_dyn} to their respective
     targets such that the agents are probabilistically
     collectively safe at all times $t$.
\end{prob}

In the statement of Problem
\ref{prob:safe_stoch_multi_agent_planning}, we specified
a risk bound for every time step ($\alpha_{i,j,t},
\beta_{i,i',t}, \kappa_{i,t}$).  On the other hand, when
a risk bound for the entire planned trajectory (e.g.
$\alpha_{i,j}$) is given over a planning horizon $T\in\bbn$, one can arrive at
$\alpha_{i,j,t}$ via \emph{risk
allocation}~\cite{blackmore2011chance} --- divide 
the risk equally across time steps with $\alpha_{i,j,t}
= \alpha_{i,j}/\horizon$ for every $t \in
\Nint{1}{\horizon}$.
\begin{figure}[t]
    \centering
    \includegraphics[width=0.98\linewidth]{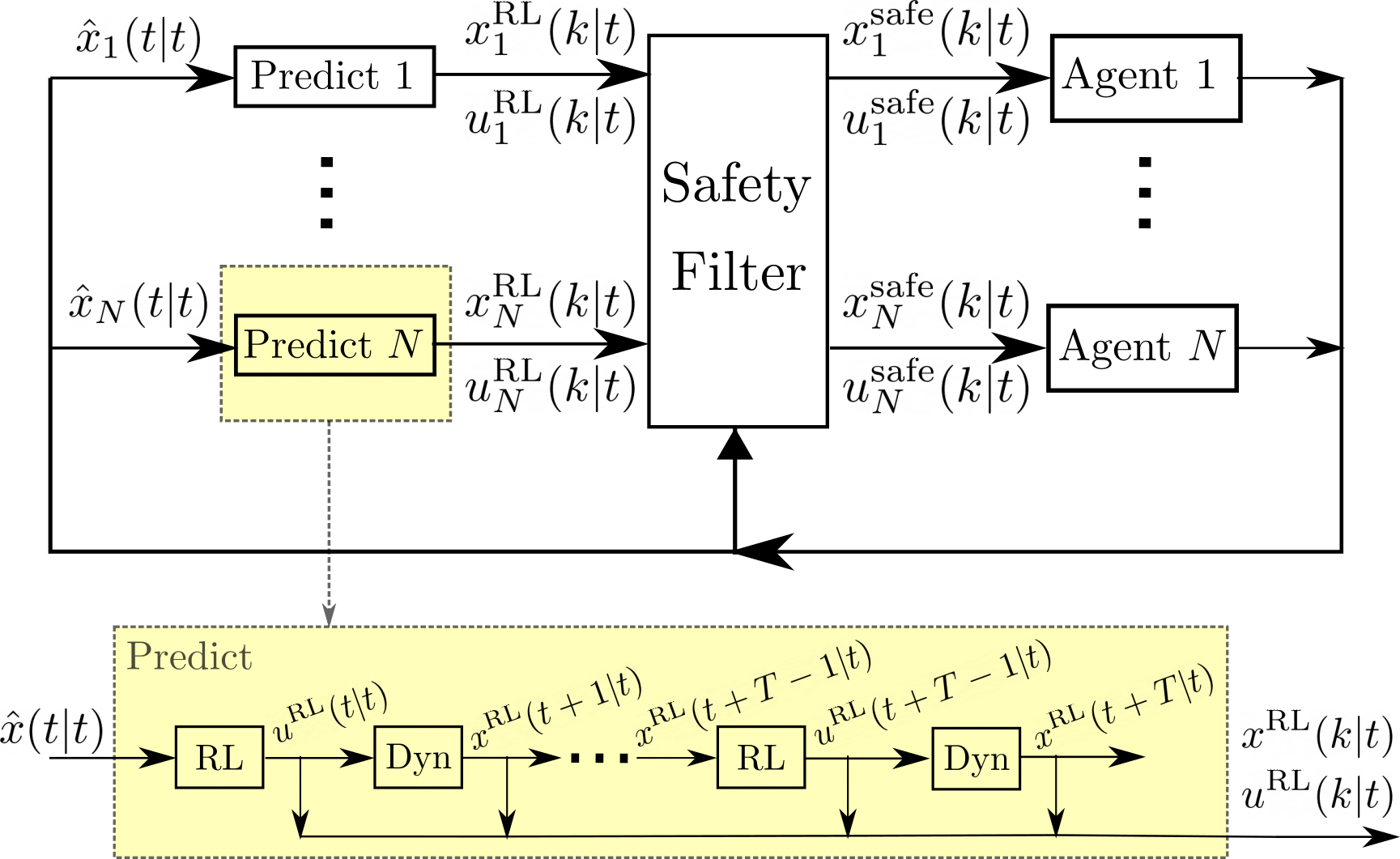} 
    \caption{The proposed solution combines single-agent
        RL-based motion planning with a constrained-control-based safety filter for
        safe multi-agent motion planning. It computes a sequence of RL
        states and controls for the horizon {$\horizon$} using a predict block. 
        Next, it uses a safety filter to render these controls safe for each agent. The predict block uses a policy network (trained offline) and the nominal dynamics \eqref{eq:mean_dynamics} to compute the RL
        controls and states.}
        \label{fig:safety_filter}
\end{figure}

\section{Proposed Solution} \label{sec:prob_sol}
We solve Problem~\ref{prob:safe_stoch_multi_agent_planning} by the following two steps.
\begin{enumerate}
    \item \emph{RL training (Offline):} We train a neural
        network
        to drive a \emph{single} agent with \emph{nominal} (deterministic) dynamics \eqref{eq:mean_dynamics} from any initial state in
        the workspace to a
        final desired state. The resulting policy learns to
        perform static obstacle collision avoidance and remain 
        within the workspace while transferring a
        single agent from its initial state to the final
        state. 
    \item \emph{Safety Filter (Online):} We use a
        constrained-control-based safety filter that uses an online
        evaluation of the RL-based motion planner. The
        safety filter suitably modifies the motion plan to
        enforce probabilistic collective safety at all time
        steps. The safety filter solves a real-time
        implementable, convex, quadratic program to determine
        the modifications.
\end{enumerate}

Figure~\ref{fig:safety_filter} depicts the proposed
solution. In the following, we describe the RL-based motion
planner, provide details of constructing the safety filter to
enforce probabilistic collective safety,  and discuss
various aspects of the proposed solution.

\subsection{Reinforcement learning-based single-agent motion planner} \label{sec:rl_train}

We design a RL-based motion planner that drives the agent
with nominal dynamics
\eqref{eq:mean_dynamics} to a specified target position $q
\in \bbr^d$ in presence of $\numObst$ static obstacles located at nominal
positions $ \mean{c}_j  \ \forall j \in
\Nint{1}{\numObst}$. 
{Here, we train a \emph{single-agent RL-based} planner in an environment devoid of other agents. After briefly discussing the motivations for such an approximation, we set up the Markov decision process used for training and characterize the neural policy obtained via single-agent RL training.}

{
The advantage of using a single-agent RL-based planner instead of the full multi-agent RL-based planner is in the ease of training. 
Specifically, a single-agent RL-based planner avoids some issues of multi-agent RL such as non-stationarity, scalability, and the diminished ability to accommodate potential changes in team size post training.
Recall that in multi-agent RL, all agents learn concurrently and thus an action taken by an individual agent affects both the reward of the other agents and the evolution of the state of the system. 
From the agent's perspective the environment is non-stationary~\cite{zhang2021multi}.
By approximating the problem and eliminating the other agents, training a single-agent RL is a stationary problem which is key for convergence results of RL training and for the reduced training effort~\cite{zhang2021multi}. 
Moreover, compared to multi-agent RL training, whose joint state space
and joint action space grow rapidly in dimension with the number of agents, the state space and the action space dimensions are fixed and independent of the team size in the single-agent RL training. 
Finally, if the team size changes post training, the proposed single-agent RL-based planner in Figure~\ref{fig:safety_filter} can still be used without any modifications as compared to a complete multi-agent RL-based planner, which may require re-training to handle changes in the team size.}

Consider a feedforward-feedback controller $\pi:\bbr^{n}\times \bbr^d \to \bbr^m$ with
\begin{align}
    u = \pi(x, r) = Kx + Fr, \label{eq:rl_control}
\end{align}
where $K$ is a stabilizing gain matrix, $F \in \bbr^{m
\times d}$ provides closed loop unitary gain with the nominal dynamics \eqref{eq:mean_dynamics}, i.e., $C(I-(A+BK))^{-1}BF = I_d$, and $r\in\bbr^d$
is the reference position command. 
We obtain the following stabilized, nominal, prediction model for the agent at any time $k\geq t$,
\begin{align}
    \mean{x}(k+1|t) &= (A+BK)\mean{x}(k|t)+BFr(k|t), \label{eq:stabilized_dyn} 
\end{align}
by closing the loop of the dynamics \eqref{eq:mean_dynamics} with the controller \eqref{eq:rl_control}.
For the measurement model \eqref{eq:measurement_model}, the predicted measurements are
$\estrealize{y}(k|t)=\mean{x}(k|t)$ for every $k\geq t$. 
By construction, the mean predicted position $\mean{p}(k|t)\to r$ as $k\to \infty$ for a constant reference command $r(k|t)=r$.

We use the following Markov decision process for training:
\begin{itemize}
    \item \emph{Observation space}: 
        We define the observation vector $o \in
        \mathbb{R}^{n + d+\numObst d}$ as the concatenated
        vector containing the current measurement of the agent $\estrealize{y} \in \mathbb{R}^n$, the
        displacement of the agent's current measured position to
        the target $(p - q) \in \mathbb{R}^d$ and
        to the $\numObst$ static obstacles $(p - \mean{c}_j)\in
        \mathbb{R}^d$, for all $j\in \Nint{1}{\numObst}$, where $p=C\hat{y}$.
    \item \emph{Action space}: The action $a\in \scra
        \subset \mathbb{R}^d$ determines the reference
        position as a perturbation $a$ to the
        target $q$, $r= q + a$. The set
        $\scra$ is compact.
    \item \emph{Step function}: The next predicted measurement\linebreak 
        $\estrealize{y}(t+1|t)=\mean{x}(t+1|t)$ is given by \eqref{eq:stabilized_dyn}.
    \item \emph{Reward function}: The instantaneous reward
        function is
        \begin{align}
            R(o) &= \zeta_\text{obs}
            \sum_{j=1}^{\numObst}\frac{1}{{\left\|
            p - \mean{c}_j\right\|}^2 -
        \gamma_j^2} + \zeta_\text{tgt} {{\left\| p - q\right\|}},\label{eq:inst_reward}
        \end{align}
    with reward parameters $\zeta_\text{obs},\zeta_\text{tgt}\leq 0$ and $\gamma_j \geq 0$. 
    Here, $\gamma_j$ is the radius of the smallest volume
    ball that covers the set $\obst{j} \oplus (-\agent)$ for each $j\in \Nint{1}{\numObst}$.
    {
    We terminate an episode when
    the agent either reaches the target or violates the \emph{nominal single-agent} safety conditions, namely the static obstacle avoidance and keep-in constraints. These constraint violations are given by:
    \begin{align*}
        (\mean{p}_i(t) \oplus \agent)\cap (\mean{c}_j(t) \oplus \obst{j}) &\neq \emptyset\text{ and }\mean{p}_i(t) \oplus \agent \nsubseteq \envPoly.
    \end{align*}}%
    When the episode terminates, we add a terminal reward or penalty as follows:
        \begin{align}
            R(o_\infty) = \begin{cases}
                \begin{array}{ll}
                    R_\text{target}, &\hspace*{-0.5em}\begin{array}{l} \text{if $\|p_\infty - q\| \leq d$}\\
                    \end{array},\\
                    P_\text{keep-in}, & \text{if $p_\infty\oplus \agent \not\subseteq \envPoly$},\\     
                    P_\text{obstacle}, & \text{if agent hits an obstacle},\\        
                \end{array}
            \end{cases}  \nonumber
        \end{align}
    with $o_\infty$ and $p_\infty$ denoting as the observation and position vectors upon termination respectively, $R_\text{target} \geq 0$, and $P_\text{keep-in}, P_\text{obstacle}\leq 0$.
\end{itemize}

We have {set up} the Markov decision process to consider deterministic nominal dynamics \eqref{eq:mean_dynamics} instead of
the original stochastic dynamics \eqref{eq:stoch_dt_time_dyn} in order to simplify the RL training. Our numerical and
hardware experiments show that the restriction to deterministic nominal dynamics does not affect the proposed solution
severely. 

\begin{rem}
    Our approach can also accommodate a known, time-varying target and biased measurement models $\mean{\rv{\eta}}\neq
    0$. We have considered a time-invariant target $q$ and an unbiased measurement model here to simplify the
    presentation. Additionally, we use the minimum volume balls with radius $\gamma_j$ in \eqref{eq:inst_reward} instead
    of $\obst{j}\oplus(-\agent)$ to simplify the collision detection while training the RL-based motion planner.  
\end{rem}

Upon completion of training, most of the existing RL
algorithms return a policy $\nnet:\mathbb{R}^{n + d+\numObst
d}\to \scra$ that provides the action to apply given an observation vector, e.g., by a neural network~\cite{raffin2019stable}.
Additionally, we can ``rollout'' the policy network $\nnet{}$ to obtain a trajectory based on the RL motion planner for a
planning horizon $T\in\mathbb{N}$. Consider any
agent $i\in\Nint{1}{\numAgents}$ that starts with the measurement $\estrealize{y}_i(t)$. 
We compute the RL motion plan $\rlmotionplan$, where
$x^{\text{RL}}_i(t|t) = \estrealize{y}_i(t)$, by
alternating between finding the control $\uirl(k|t)$ given the predicted RL state $x^{\text{RL}}_i(k|t)$ and predicted observation vector $o_i(k|t)$ at some time $k\geq t$ using $\pi$,
\begin{align}
    u_i^{\text{RL}}(k|t) &= \pi\left({x^{\text{RL}}_i(k|t), q+\nnet(o_i(k|t)}\right),\label{eq:rl_control_predict}
\end{align}
and predicting the next RL state $x^{\text{RL}}_i(k+1|t)$ using \eqref{eq:stabilized_dyn} and the corresponding predicted observation vector $o_i(k+1|t)$.

The generated motion plan $\rlmotionplan$ does not satisfy
probabilistic collective safety, since RL cannot
guarantee collision-free trajectories (it only penalizes
collisions and is subject to training errors), and the generated RL motion plan completely
ignores inter-agent collision avoidance and the effect of the process and measurement noises.

\subsection{Safety Filter} \label{sec:safety_filter}

We now generate corrections to the RL-based motion plan $\rlmotionplan$ using a constrained-control-based safety filter that ensures the satisfaction of probabilistic collective safety at all times. Consider the following optimization problem with the stochastic information,
\begingroup
    \makeatletter\def\f@size{9}\check@mathfonts
    \begin{subequations}
        \begin{align}
            \hspace*{-1em}\underset{\{\Uisafe(t)\}_{i=1}^N}{\mathrm{min}}
                &\ \sum_{k\in\Nint{t}{t+T-1}}\sum_{i\in\Nint{1}{N}}
                \lambda_{i,k}{\|\uirl(k|t) - \uisafe(k|t) \|}^2\label{eq:optimization_centralized_cost}\\
            \mathrm{s.t.} 
                &\ \text{Dynamics
                    \eqref{eq:stoch_dt_time_dyn} and
                    \eqref{eq:pos} with $u=\usafe$,}\\
                &\ \rv{x}_i(t|t) = \estrealize{y}_i(t) - \rv{\eta}(t),\hspace*{4.9em} \forall i \in \Nint{1}{\numAgents}, \label{eq:optimization_centralized_agent_init_state}\\
                &\ \uisafe(k|t) \in \inputSet,\hspace*{1.05em}\forall k\in\Nint{t}{t+T-1},\ \forall i \in \Nint{1}{\numAgents},\label{eq:optimization_centralized_control}\\
                &\  
                \text{Probabilistic collective safety
                 at $k$},\hspace*{0.75em} \forall k\in\Nint{t}{t+T}, 
                \label{eq:prob_collective_safety} \\
                &\  
                \text{Terminal constraints for recursive feasibility}
                \label{eq:recursive_feasbility_condition},
        \end{align}\label{prob:safety_filter}%
    \end{subequations}%
\endgroup
where $\Uisafe(t)=\alluisafe$ for each $i\in\Nint{1}{N}$, and $\lambda_{i,k}\geq 0$ are pre-specified weights on
the deviations ${\|\uirl(k|t) - \uisafe(k|t) \|}^2$ for $i\in\Nint{1}{N}$ and $k\in\Nint{t}{t+T-1}$. 

\begin{figure*}[t]
\setcounter{equation}{12}
\begin{subequations}
\begin{align}
    \forall j\in\Nint{1}{N_O},&&\zijobs \cdot (\mean{p}_i(k|t) - \mean{c}_j)&\geq S_{\obst{j}}(\zijobs) + S_{(-\agent)}(\zijobs) - \|(\Sigma_{p_i}(k|t) +
    \Sigma_{c_j})^{1/2} {\zijobs} \| \Phiinv(\alpha_{i,j,t}),\label{eq:prop_1_eqns_obs}\\
    \forall j\in\Nint{1}{i-1},&&\zijagent \cdot (\mean{p}_i(k|t) - \mean{p}_{j}(k|t)) &\geq  S_{\agent}(\zijagent) + S_{(-\agent)}(\zijagent)  - \| (\Sigma_{p_i}(k|t) +
    \Sigma_{p_{j}}(k|t))^{1/2} {\zijagent} \| \Phiinv(\beta_{i,j,t}),\label{eq:prop_1_eqns_agent}\\
    \forall j\in\Nint{1}{\numEnvHalfspaces},&&\envHalfspaceA_j \cdot \mean{p}_i(k|t) &\leq \envHalfspaceB_j - 
    S_{\agent}(\envHalfspaceA_j) - \| \Sigma^{1/2}_{p_i}(k|t)
    \envHalfspaceA_j \| \Phiinv\left({1-(\kappa_{i,t}/\numEnvHalfspaces)}\right).\label{eq:prop_1_eqns_env}
\end{align}\label{eq:prop_1_eqns}%
\end{subequations}%
\vspace*{-2em}
\par\noindent\rule{\textwidth}{0.4pt}
\setcounter{equation}{13}
\end{figure*}

The safety filter \eqref{prob:safety_filter} takes the RL control sequence $\rlcontrolseq$ that
is generated using the RL-based single-agent motion planner,
and computes safe control inputs $\alluisafe$ within the control set \eqref{eq:optimization_centralized_control}
that minimally deviate from the corresponding RL control
inputs \eqref{eq:optimization_centralized_cost}, while
enforcing probabilistic collective safety constraints
\eqref{eq:prob_collective_safety}. The constraint \eqref{eq:optimization_centralized_agent_init_state} defines the distribution of the noisy current state $\rv{x}(t|t)$ from the current measurement $\hat{y}(t)$ and the measurement model \eqref{eq:measurement_model}. Additionally, to avoid
computing control actions $u_i^{\text{safe}}$ that may
render the optimization problem
\eqref{prob:safety_filter} in the safety filter infeasible in {the} future, we include
terminal state constraints
\eqref{eq:recursive_feasbility_condition} that, when
designed as explained later, provide recursive
feasibility.  Only the first safe control
$u_i^{\text{safe}}(t|t)$ is applied for each agent $i$, 
and then \eqref{prob:safety_filter} is solved again at time $t+1$ in an
MPC-like fashion~\cite{borrelli2017predictive}. 

The safety filter \eqref{prob:safety_filter} is a nonlinear,
non-convex, and stochastic optimization problem due to
\eqref{eq:prob_collective_safety} and
\eqref{eq:recursive_feasbility_condition}, and as a
consequence in general not real-time implementable.
Therefore, we reformulate \eqref{prob:safety_filter} by
convexifying the constraints and replacing the chance
constraints by deterministic risk-tightened constraints,
which we describe next.

\subsection{Convexified constraints for probabilistic collective safety}
\label{sub:reformulate_chance_const_mpc}
We now present a convex, deterministic reformulation of
\eqref{eq:prob_collective_safety} that relies on 
well-known properties of Gaussian random vectors
and the generated motion plan $\rlmotionplan$.

\begin{lemma}[\textsc{Gaussian random vectors~\cite[Sec. 4.4.2]{boyd2004convex}}] \label{lemma:gauss}
    1) Let $N_I\in\bbn$. Given $n$-dimensional Gaussian random vectors $\rv{y}_i \sim \caln(\mean{y}_i, \Sigma_{y_i})$
    with $\mean{y}_i \in \bbr^n$, $\Sigma_{y_i} \in \bbr^{n\times n}$, and matrices $Y_i\in\bbr^{m\times n}$ for each $i
    \in \Nint{1}{N_I}$, then the random vector $\rv{y} = \sum_{i=1}^{N_I} Y_i \rv{y}_i$ is also Gaussian, with
    \begin{align}
         \rv{y} \sim \caln\left(\sum\nolimits_{i=1}^{N_I} Y_i \mean{y}_i, \sum\nolimits_{i=1}^{N_I} Y_i \Sigma_{y_i} Y_i^\top\right). \label{eq:gauss_lin_comb}
    \end{align}
    \\2) Given $\rv{y} \sim \caln(\mean{y}, \Sigma_y)$ with
    $\mean{y} \in \bbr^n, \Sigma_y \in \bbr^{n \times n}$,
    $a \in \bbr^n$, $b \in \bbr$, and the risk bound
    $\alpha$, then
    \begingroup
        \makeatletter\def\f@size{9.5}\check@mathfonts
    \begin{subequations}
    \begin{align}
        \bbp(a \cdot \rv{y} \leq b) \leq \alpha &\iff 
        a\cdot \mean{y} \geq b - \| \Sigma_y^{1/2} a \| \Phiinv(\alpha), \label{eq:simple_cc_reform1} \\
        \bbp(a \cdot \rv{y} \geq b) \leq \alpha &\iff 
        a \cdot \mean{y} \leq b - \| \Sigma_y^{1/2} a \| \Phiinv(1-\alpha), \label{eq:simple_cc_reform2}
    \end{align}
    \end{subequations}
    \endgroup
    where $\Phiinv$ is the inverse cumulative distribution
    function of a standard Gaussian random variable.
\end{lemma}
\begin{figure*}
    \centering
    \includegraphics[height=0.13\textheight]{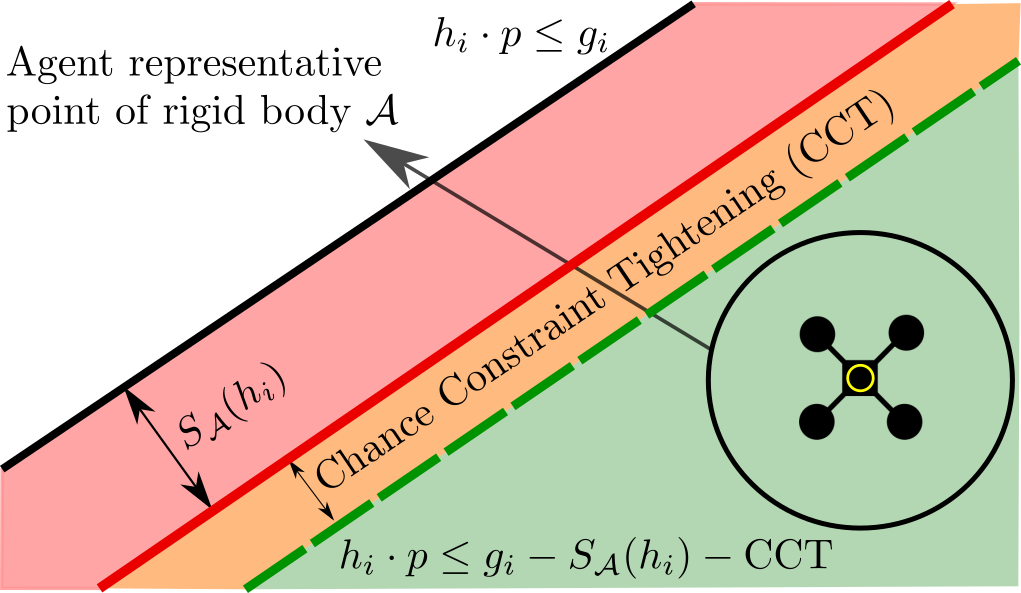} \hspace*{6em}
    \includegraphics[trim={15 10 0 0}, clip, height=0.13\textheight]{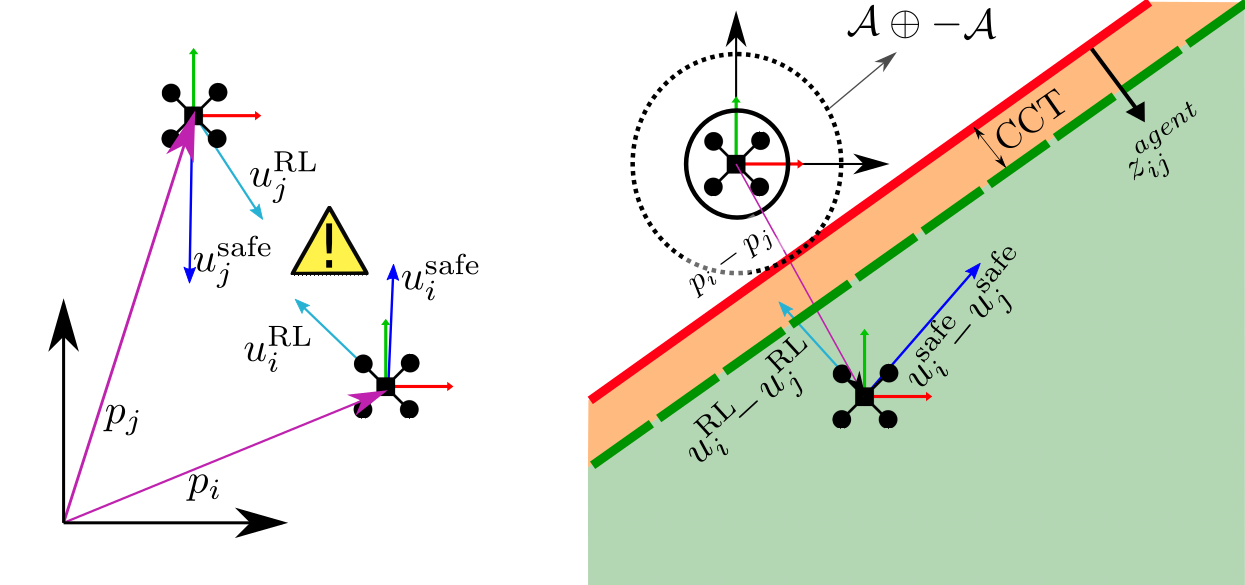}
    \caption{Probabilistic collective safety constraints (Definition~\ref{def:prob_collective_safety}) enforced as linear constraints --- (Left) Keep-in constraint (black) tightened by the support of $\cala$ (red). (Right) Inter-agent collision avoidance constraint uses the support of $\cala \oplus - \cala$ (red). Both red constraints are tightened by the chance constraint term resulting in a new constraint (dashed green).
    }
    \label{fig:constraints}
\end{figure*}

From \eqref{eq:optimization_centralized_agent_init_state}
and \eqref{eq:measurement_model}, $\rv{x}(t)\sim
\mathcal{N}(\estrealize{y}_i(t), \Sigma_{\eta})$. 
For every agent $i\in\Nint{1}{\numAgents}$, the predicted state and position at any time $k> t$, 
\begin{subequations}
\allowdisplaybreaks
\begin{align}
    \rv{x}_i(k|t) &\sim \caln (\mean{x}_i(k|t),
    \Sigma_{x_i}(k|t)),\\
    \rv{p}_i(k|t) &\sim \caln (\mean{p}_i(k|t),
    \Sigma_{p_i}(k|t)),\\
    \mean{x}_i(k|t) &= A^{k-t}\estrealize{y}_i(t) +
    \sum_{j=t}^{k-1}A^{k-(j+1)} B \uisafe(j|t),\label{eq:pred_state_mean}\\
    \mean{p}_i(k|t) &= C \mean{x}_i(k|t),\label{eq:pred_position_mean} \\
    \hspace*{-0.4em}\Sigma_{x_i}(k+1|t) &= A\Sigma_{x_i}(k|t)A^\top +
    \Sigma_w,\label{eq:pred_state_cov}\\
    \Sigma_{p_i}(k|t) &= C \Sigma_{x_i}(k|t) C^\top.\label{eq:pred_position_cov}
\end{align}\label{eq:pred_stoch}%
\end{subequations}%
using the stochastic dynamics \eqref{eq:stoch_dt_time_dyn}, \eqref{eq:optimization_centralized_agent_init_state}, and
\eqref{eq:simple_cc_reform1} in Lemma~\ref{lemma:gauss}.  We observe that $\mean{x}_i(k|t)$ and $\mean{p}_i(k|t)$ depend on the decision variables
$\uisafe$, but $\Sigma_{x_i}(k|t)$ and $\Sigma_{p_i}(k|t)$ do not. Thus, $\Sigma_{x_i}(k|t)$ and $\Sigma_{p_i}(k|t)$ may
be computed offline.

\begin{prop}[\textsc{Risk-Tightened Sufficient Safety Constraints}] \label{prop:risk_tightened_constraints}
Given a polytope\linebreak $\envPoly = \cap_{i\in
\Nint{1}{\numEnvHalfspaces}}\left\{{p \in
\bbr^d:\envHalfspaceA_i \cdot p \leq
\envHalfspaceB_i}\right\}$ with $\numEnvHalfspaces \in \bbn$
halfspaces characterized by
$\{\envHalfspaceA_i,\envHalfspaceB_i\}_{i=1}^{\numEnvHalfspaces}$, $\envHalfspaceA_i \in \bbr^d$ and $\envHalfspaceB_i \in
\bbr$, and user-defined unit vectors $\zijobs, \zijagent
\in \bbr^{d}$. Then, for every $i \in \Nint{1}{N}$ and $k\in\Nint{t}{t+\horizon}$, \eqref{eq:prop_1_eqns} is sufficient for \eqref{eq:prob_collect_static_obstacle_cc},
\eqref{eq:prob_collect_inter_agent_cc}, and
\eqref{eq:prob_collect_keepin_cc} to hold.
\setcounter{equation}{16}
\end{prop}
We provide the proof of Proposition~\ref{prop:risk_tightened_constraints} in Appendix~\ref{app:risk_tightened_constraints_proof}. 

The reformulation in
Proposition~\ref{prop:risk_tightened_constraints} follows
from applying computational geometry arguments to convexify the chance
constraints \eqref{eq:prob_collect_static_obstacle_cc}--\eqref{eq:prob_collect_keepin_cc} using supporting
hyperplanes defined by user-specified vectors $\zijobs, \zijagent
$, and then applying Lemma~\ref{lemma:gauss} and Boole's law to arrive at 
\eqref{eq:prop_1_eqns}.
From \eqref{eq:pred_state_mean} and \eqref{eq:pred_position_mean}, the constraints in Proposition~\ref{prop:risk_tightened_constraints} are linear
inequalities in the decision variables $\alluisafe$ for every $i \in \Nint{1}{\numAgents}$. 
Figure~\ref{fig:constraints} illustrates the reformulated constraints of Proposition~\ref{prop:risk_tightened_constraints}.

\subsection{Ensuring recursive feasibility using reachability} \label{sub:recursive_feasibility_reform}

We now turn our attention to
\eqref{eq:recursive_feasbility_condition} that is designed
to ensure that
\eqref{prob:safety_filter} remains feasible in subsequent control time steps. 
For recursive feasibility
\eqref{eq:recursive_feasbility_condition}, we enforce the
existence of a terminal set and a control
$u_i^\text{recurse}(k)\in\inputSet$ for all $k\geq t+\horizon$ for each agent $i$ such
that the following constraints hold for all $k\geq t+\horizon$,
\begin{subequations}
\begin{align}
    {\small\text{$\bbp((\rv{p}_i(k|t) \oplus \agent)\cap (\rv{c}_j \oplus \obst{j}) \neq \emptyset)$}} &\leq \delta, \label{eq:recurse_prob_collect_static_obstacle_cc}\\
    {\small\text{$\bbp((\rv{p}_i(k|t) \oplus \agent)\cap (\rv{p}_j(k|t) \oplus \agent) \neq \emptyset) $}}&\leq \delta, \label{eq:recurse_prob_collect_inter_agent_cc},\\
    {\small\text{$\bbp((\rv{p}_i(k|t) \oplus \agent) \not\subseteq \envPoly)$}}&\leq \delta, \label{eq:recurse_prob_collect_keep_in_cc}
\end{align}\label{eq:recurse_conditions_cc}%
\end{subequations}%
where $\delta\in(0,1)$ is a (small) user-specified risk threshold.

Existing literature in constrained control typically enforces recursive feasibility
using control invariant or
positive invariant sets~\cite{mesbah2016stochastic,borrelli2017predictive}. 
However, characterization of such sets can be challenging in
our setting due to the inherent non-convexity of the
probabilistic collective safety constraints. Alternatively,
one can approximately enforce these constraints by
truncating the recursive feasibility criterion to a finite
but long horizon, and then utilizing stochastic reachability~\cite{SummersLygeros,VinodAutomatica,malone2017hybrid}.

For the sake of tractability, we enforce
\eqref{eq:recurse_conditions_cc} approximately by imposing
chance constraints on the terminal states, while ignoring
the stochasticity in the future time steps. We characterize
these constraints using appropriately defined \emph{avoid
sets} (also known as \emph{inevitable collision
states}~\cite{malone2017hybrid} or \emph{capture
sets}~\cite{aubin2011viability}) and \emph{viability sets}
(also known as \emph{controlled invariant
sets})~\cite{borrelli2017predictive,maidens2013lagrangian}.

\begin{defn}[\textsc{Avoid set and viability set}~\cite{borrelli2017predictive}]\label{def:avoid_set_defn}
For a (bad) set $\mathscr{B}\subset \bbr^n$, linear dynamics \eqref{eq:mean_dynamics}, and a control constraint set $\mathcal{U}$, we define an avoid set as follows,
\begin{align}
    \AvoidSet(\mathscr{B})=\left\{\bar{x}(0) \middle|\begin{array}{c}
    \exists t\in \bbn,\ \forall u(t)\in\mathcal{U},\\
    \bar{x}(t+1)=A \bar{x}(t) + Bu(t) \in \mathscr{B}
    \end{array}\right\}\nonumber 
\end{align}
For a (good) set $\mathscr{G}\subset \bbr^n$, we define a viability set as follows,
\begingroup
    \makeatletter\def\f@size{9.5}\check@mathfonts
\begin{align}
    \ViabSet(\mathscr{G})=\left\{\bar{x}(0) \middle|\begin{array}{c}
    \forall t\in \bbn,\ \exists u(t)\in\mathcal{U},\\
    \bar{x}(t+1)=A \bar{x}(t) + Bu(t) \in \mathscr{G}
    \end{array}\right\}.\nonumber
\end{align}
\endgroup
By construction, 
we have
\begin{align}
\bbr^n\setminus\ViabSet(\mathscr{G})=\AvoidSet(\bbr^n\setminus\mathscr{G}).\label{eq:avoid_set_viab_set}
\end{align}
\end{defn}
Informally, $\AvoidSet(\mathscr{B})$ is the set of mean
initial states from which the mean trajectory of \eqref{eq:mean_dynamics} enters
the bad set $\mathscr{B}$ at some time $t$,
irrespective of the control choices. On
the other hand, $\ViabSet(\mathscr{G})$ is the set of mean
initial states from which the mean trajectory of \eqref{eq:mean_dynamics} remains within the good set $\mathscr{G}$ for all time $t$,
by some appropriate choice of control actions. 

For any time $t$, assume that the agents evolve by stochastic dynamics
\eqref{eq:stoch_dt_time_dyn} with imperfect
measurements according to \eqref{eq:measurement_model} during the planning interval ($k\in\Nint{t}{t+\horizon-1}$), and they evolve by nominal dynamics \eqref{eq:mean_dynamics} with perfect measurements beyond the planning horizon
($k\geq t+\horizon$). Since the safety filter solves \eqref{prob:safety_filter} at every $t$ based on the new
measurement, the impact of this assumption is mild for
sufficiently long planning horizon $T$. Under this assumption, we
construct the following approximation of
\eqref{eq:recurse_conditions_cc} using Definition~\ref{def:avoid_set_defn},
\begin{subequations}
\begin{align}
    {\small\text{$\bbp((\rv{x}_i(k+\horizon|k) - \rv{c}^{\text{lift}}_j(k+\horizon|k)) \in\AvoidSet(\obst{j} \oplus (-\agent))) $}} &\leq \delta, \label{eq:avoid_recurse_prob_collect_static_obstacle_cc}\\
    {\small\text{$\bbp((\rv{x}_i(k+\horizon|k) - \rv{x}_j(k+\horizon|k))\in\AvoidSet(\agent \oplus (-\agent))) $}}&\leq \delta,\label{eq:avoid_recurse_prob_collect_inter_agent_cc}\\
    {\small\text{$\bbp(\rv{x}_i(k+\horizon|k)
\not\in\ViabSet(\envPoly\ominus \agent))$}}&\leq \delta, \label{eq:avoid_recurse_prob_collect_keep_in_cc}
\end{align}\label{eq:avoid_recurse_conditions_cc}%
\end{subequations}%
where $\rv{c}^{\text{lift}}_j\in\bbr^n$ is obtained by lifting the position to $\bbr^n$ with added components set to zero, since
the obstacles are static. \eqref{eq:avoid_recurse_prob_collect_keep_in_cc} uses
\eqref{eq:avoid_set_viab_set} for ease in implementation.
Informally,
\eqref{eq:avoid_recurse_prob_collect_static_obstacle_cc} and
\eqref{eq:avoid_recurse_prob_collect_inter_agent_cc} require
the agents to be in configurations that lead to
collision with static obstacles or each other with at most a probability of $\delta$, and
\eqref{eq:avoid_recurse_prob_collect_keep_in_cc} requires
the probability that the agents are in configurations from which they can not remain within the workspace is at most $\delta$.

\begin{figure*}[t]
\begin{subequations}
\begin{align}
    \forall j\in\Nint{1}{N_O},&&\ellijobs \cdot (\mean{x}_i(t+\horizon|t)  -\mean{c}_j^{\text{lift}}) &\geq S_{\ellipAvoid{\obst{j}}}(\ellijobs)- \|
            (\Sigma_{x_i}(t+\horizon|t) +
            \Sigma_{c^{\text{lift}}_j})^{1/2} {\ellijobs}\|\Phiinv(\delta),\\ 
    \forall j\in\Nint{1}{i-1},&&\ellijagent \cdot (\mean{x}_i(t+\horizon|t) - \mean{x}_{j}(t+\horizon|t)) &\geq S_{\ellipAvoid{\agent}} - \| (\Sigma_{x_i}(t+\horizon|t)
            + \Sigma_{x_{j}}(t+\horizon|t))^{1/2}
            {\ellijagent} \| \Phiinv(\delta),\\
            \forall j \in \Nint{1}{\numPolyViabHalfspaces},&&\polyViabHalfspaceA_j  \cdot \mean{x}_i(t+\horizon|t) &\leq \polyViabHalfspaceB_j - \|
            \Sigma_{x_i}^{1/2}(t+\horizon|t)
            \polyViabHalfspaceA_j \| \Phiinv\left({1-(\delta/\numPolyViabHalfspaces)}\right).
\end{align}\label{eq:prop_2_eqns}%
\end{subequations}%
\vspace*{-2em}
\par\noindent\rule{\textwidth}{0.4pt}
\end{figure*}

The constraints \eqref{eq:avoid_recurse_conditions_cc} are
tractable when the sets
$\AvoidSet(\agent \oplus (-\agent))$, $\AvoidSet(\agent
\oplus (-\obst{j}))$, and $\ViabSet(\envPoly\ominus \agent)$ are
convex.
Recall that $\AvoidSet(\mathcal{B})$ is typically
non-convex, even when $\mathscr{B}\in\{\agent \oplus
(-\agent),\agent
\oplus (-\obst{j})\}$ is a convex
polytope~\cite{borrelli2017predictive}. This complicates the
enforcement of
\eqref{eq:avoid_recurse_prob_collect_static_obstacle_cc} and
\eqref{eq:avoid_recurse_prob_collect_inter_agent_cc}. For
the sake of tractability and ensuring conservativeness, we propose
Algorithm~\ref{algo:det_avoid_set} to compute an ellipsoidal
outer-approximation of $\AvoidSet(\mathcal{B})$. Outer-approximations of $\AvoidSet$ are also sufficient to enforce \eqref{eq:avoid_recurse_prob_collect_static_obstacle_cc} and \eqref{eq:avoid_recurse_prob_collect_inter_agent_cc}. 
On the other hand, 
Algorithm~\ref{algo:viab_set} provides an exact approach to
compute $\ViabSet(\envPoly\ominus \agent)$ for convex and
compact polytopes $\envPoly$ and $\agent$. All operations in Algorithms~\ref{algo:det_avoid_set} and~\ref{algo:viab_set} can be easily accomplished using computational geometry tools and convex optimization, see~\cite{boyd2004convex,maidens2013lagrangian,borrelli2017predictive,MPT3} for more details.

\begin{algorithm}[t]
\caption{Computation of $\AvoidSetP(\mathscr{B})$ (See~\cite[Sec. 10.2]{borrelli2017predictive} for recursion)}\label{algo:det_avoid_set}
\newcommand{\ListSets}{\textit{ListOfSets}}
\newcommand{\CurrentAvoidSet}{\textit{CurrentSet}}
\newcommand{\ConvHull}{\textit{ConvexHullOfList}}
\begin{algorithmic}[1]
    \Require Linear dynamics \eqref{eq:mean_dynamics}, control constraint set $\inputSet$,\linebreak convex and compact polytope $\mathscr{B}$.

    \Ensure $\AvoidSetP(\mathscr{B})$.
    
    \State $\ListSets \gets  [\mathscr{B}]$,  $\CurrentAvoidSet\gets \mathscr{B}$
    
    \State \textbf{while} $\CurrentAvoidSet$ is non-empty
    
    \State\hspace*{1em} $\CurrentAvoidSet\gets A^{-1}(\CurrentAvoidSet\ominus B\mathcal{U})$
    
    \State\hspace*{1em} Append $\CurrentAvoidSet$ to $\ListSets$
  
    \State $\ConvHull\gets$ convex hull of 
    $\ListSets$  
    
    \State $\AvoidSetP(\mathscr{B}) \gets$ minimum volume ellipsoid \linebreak containing $\ConvHull$ (see~\cite[Sec. 8.4.1]{boyd2004convex})
  \end{algorithmic}
\end{algorithm}

\begin{algorithm}[t]
\caption{Computation of $\ViabSet(\mathscr{G})$~\cite{maidens2013lagrangian}}\label{algo:viab_set}
\newcommand{\PrevViabSet}{\textit{PrevSet}}
\newcommand{\CurrentViabSet}{\textit{CurrentSet}}
\begin{algorithmic}[1]
    \Require Linear dynamics \eqref{eq:mean_dynamics}, control constraint set $\inputSet$, \linebreak convex and compact polytope $\mathscr{G}$.

    \Ensure $\ViabSet(\mathscr{G})$.
    
    \State $\CurrentViabSet \gets  \mathscr{G}$, $\PrevViabSet\gets \emptyset$
    
    \State \textbf{while} $\CurrentViabSet$ is not equal to $\PrevViabSet$
    
    \State\hspace*{1em} $\PrevViabSet\gets\CurrentViabSet$
    
    \State\hspace*{1em} $\CurrentViabSet\gets \mathscr{G}\cap A^{-1}(\CurrentViabSet\oplus (- B\mathcal{U}))$
  
    \State $\ViabSet(\mathscr{G})\gets \CurrentViabSet$
  \end{algorithmic}
\end{algorithm}
We conclude this section by characterizing a set of linear
constraints that are sufficient to enforce
\eqref{eq:avoid_recurse_conditions_cc}. The proof of
Proposition~\ref{prop:risk_tightened_terminal_constraints}
uses the same arguments as that seen in  
Proposition~\ref{prop:risk_tightened_constraints}. 

\begin{prop}[\textsc{Risk-Tightened Sufficient Terminal Recursive Feasibility Constraints}] \label{prop:risk_tightened_terminal_constraints}
    Given a collection of user-defined unit vectors $\ellijobs, \ellijagent \in \bbr^{n}$, let $\ellipAvoid{\obst{j}} \triangleq
    \AvoidSetP(\obst{j} \oplus (-\agent))$ and
    $\ellipAvoid{\agent}\triangleq\AvoidSetP(\agent \oplus
    (-\agent))$ denote ellipsoidal outer-approximations of the corresponding avoid sets, and
    $\polyViab\triangleq\ViabSet(\envPoly \ominus \agent)$
    denote a polytope with $\numPolyViabHalfspaces$ halfspace constraints, $\polyViab = \cap_{i\in \Nint{1}{\numPolyViabHalfspaces}}\left\{{x \in \bbr^n:\polyViabHalfspaceA_i \cdot x \leq \polyViabHalfspaceB_i}\right\}$.
    Then, for every $i \in \Nint{1}{N}$, \eqref{eq:prop_2_eqns} is sufficient for  \eqref{eq:avoid_recurse_conditions_cc} to hold.
\end{prop}

Proposition~\ref{prop:risk_tightened_terminal_constraints}
characterizes linear constraints that are
sufficient to enforce
\eqref{eq:avoid_recurse_conditions_cc}. For \eqref{eq:avoid_recurse_prob_collect_static_obstacle_cc} and
\eqref{eq:avoid_recurse_prob_collect_inter_agent_cc}, the
sufficient condition is obtained by tightening the
complement of a supporting halfspace of the convex
outer-approximations of the avoid set. For
\eqref{eq:avoid_recurse_prob_collect_keep_in_cc}, the
sufficient condition utilizes Boole's inequality and uniform
risk allocation.

\subsection{Reformulated Risk-Tightened MPC Safety Filter}
\label{sub:final_safety_filter}

Following the reformulations discussed in Sections~\ref{sub:reformulate_chance_const_mpc} and~\ref{sub:recursive_feasibility_reform}, we obtain a quadratic program \eqref{prob:safety_filter_reform},
\begingroup
    \begin{align}
        \hspace*{-2em}\begin{array}{rl}
            \underset{\{\Uisafe(t)\}_{i=1}^N}{\mathrm{minimize}}
                &\eqref{eq:optimization_centralized_cost}\\
            \mathrm{subject\ to} 
                &\eqref{eq:optimization_centralized_control},\ \eqref{eq:pred_state_mean},\  \eqref{eq:pred_position_mean},\  \eqref{eq:pred_state_cov},\ \eqref{eq:pred_position_cov},\\
                &\text{\eqref{eq:prop_1_eqns} for every $i\in\Nint{1}{N},\ k\in \Nint{t}{t+\horizon}$},\\
                &\text{\eqref{eq:prop_2_eqns} for every $i\in\Nint{1}{N}$}, 
        \end{array}\label{prob:safety_filter_reform}%
        \end{align}
where $\Uisafe(t)=\alluisafe$ for each $i\in\Nint{1}{N}$.
\eqref{prob:safety_filter_reform} uses the mean states and positions of the agents, and the deterministic linear constraints characterized in Propositions~\ref{prop:risk_tightened_constraints} and~\ref{prop:risk_tightened_terminal_constraints} for probabilistic collective safety and recursive feasibility. A solution of \eqref{prob:safety_filter_reform} is a feasible (but not necessarily optimal) solution of \eqref{prob:safety_filter}.

From the setting of \eqref{prob:safety_filter_reform}, it is
evident that the specialization of the RL-based motion
planner to the deterministic nominal dynamics
\eqref{eq:mean_dynamics} does not affect the proposed
solution adversely. Motivated by the superposition
principle, we have used a simpler RL-based motion planner that
considers deterministic dynamics \eqref{eq:mean_dynamics} instead of the stochastic dynamics \eqref{eq:stoch_dt_time_dyn},
and delegated the responsibility of probabilistic collective
safety under \eqref{eq:stoch_dt_time_dyn} to the safety
filter.

\subsection{Discussion}
{

\subsubsection{Choice of Dynamics}
    Our primary targets are quadrotors as we describe in Section \ref{sec:implementation}. 
    Waypoint tracking for quadrotors using on-board controllers is now well-known~\cite{augugliaro2012generation,preiss2017crazyswarm}. Consequently, assuming linear dynamics \eqref{eq:stoch_dt_time_dyn} is appropriate since the safety filter can generate safe waypoints that deviates minimally from the RL-based motion 
    plan. 
    
    Theoretically, it is possible to apply the proposed solution to nonlinear dynamics. However, the construction of terminal sets for collision avoidance and recursive feasibility, similar to the sets proposed in Section~\ref{sub:recursive_feasibility_reform}, become more challenging~\cite{mesbah2016stochastic}
    On the other hand, our approach achieves recursive feasibility in the presence of stochastic process noises. Using process noise to (conservatively) model the linearization error when using linear models for nonlinear dynamics, we can use the proposed approach to provide (conservative) safety guarantees.
}

{\subsubsection{Gaussian Noise Assumption}
    In} our problem statement, we assumed Gaussian noise and imposed chance constraints. Alternatively, 
    we can use other risk metrics based on axiomatic risk theory and more generalized noise distributions~\cite{majumdar2020should, safaoui2021risk, lindemann2021reactive}. However, most of these approaches either do not admit closed-form deterministic reformulations resulting in high computational costs, or are overly
    conservative. For example, our assumptions on $\rv{w}$ and $\rv{\eta}$ having a Gaussian distribution can be relaxed to
    any probability
    distribution that has a pre-specified mean and covariance. 
    In this case, the reformulated constraints are similar to \eqref{eq:prop_1_eqns} and \eqref{eq:prop_2_eqns} but
    with $\Phiinv(\alpha)$ terms replaced by the Chebychev bound
    $\sqrt{\frac{1-\alpha}{\alpha}}$~\cite{calafiore2006distributionally}. However, the resulting deterministic sufficient conditions are far more conservative than those in Propositions~\ref{prop:risk_tightened_constraints} and~\ref{prop:risk_tightened_terminal_constraints}~\cite[Fig. 2]{farina2016stochastic}.

{\subsubsection{Ellipsoidal Convex Set Usage}
    We} recommend using ellipsoidal representations (or outer-approximations) for various convex sets, primarily due to the convexification step presented in Section~\ref{sub:recursive_feasibility_reform}. Ellipsoids ${\mathcal{E}(c,Q)}=\{x|(x -c)\cdot (Q^{-1}(x-c))\leq 1\}$ admit a closed form solution for the support function $\rho_{\mathcal{E}(c,Q)}(\ell)=\ell\cdot c + \sqrt{\ell \cdot (Q \ell)}$, and the supporting hyperplane changes smoothly along the set boundary with changing $\ell$.
    Compared to that, the support function of a polytope
    requires solving a linear program, and may change abruptly when changing $\ell$.

{
\subsubsection{Safety Filtering with Other Motion Planners}
We use the proposed safety filter \eqref{prob:safety_filter} in conjunction with single-agent RL motion planning, since RL-based planners have become popular in recent literature (see discussion in  Section~\ref{sec:intro}) but lack safety guarantees especially in terms of enforcing (hard) constraints.
While the proposed combination of RL and safety filter can provide hard constraint satisfaction guarantees, the safety filter's applicability is not limited to single-agent RL-based planners.
As illustrated in Figure \ref{fig:safety_filter} and as seen from the derivations, the safety filter only requires the agents' reference state and control trajectories.
These inputs may also be obtained from many other planners, including more traditional ones such as sampling-based. 
For example, RRT-based planners~\cite{lavalle1998rapidly, karaman2011sampling} can be used for single-agent motion planning while avoiding static obstacles in the environment. The multi-agent plans can then be obtained by combining separate single-agent RRT-based plans using the proposed safety filter to guarantee inter-agent collision avoidance.
This allows more efficient computations and memory reduction with respect to applying sampling-based planning to the multi-agent problem due to the smaller dimension and reduced number of collisions to be checked.
}

\newcommand{\Nrl}{N_\text{few}}
\subsubsection{Intermediate multi-agent RL-based planners}
The proposed approach can also be applied to the intermediate case of a planner for multiple agents $\Nrl$, but less than the total number $\numAgents$.
In this case, multiple planners generate plans each for $\Nrl$ agents up to the total number $\numAgents$. Each group of $\Nrl$ may be collision-free, but the safety filter is applied to ensure safety between agents in different groups.
Overall, the fundamental idea behind our approach is to take a challenging motion planning problem, approximate it by a problem that is significantly simpler to solve, at the price of losing safety due to the approximation, and then recovering it by the safety filter.
In the case of multi-agent planning, approximation is done by reducing the amount of agents, hence here we discussed the largest possible reduction that provides the largest simplification, that is only one agent is considered in planning, but the approach will also work for any intermediate case.

\section{Implementation Details and Experiment Setup} \label{sec:implementation}
\textbf{Dynamics:}
We used the Crazyflie 2.1 quadrotors~\cite{crazyflie} as our target platform. We flew all the quadrotors at the same
height of $0.95$ m to make the collision avoidance problem
more challenging. While it would be possible to resolve collisions by flying the drones at different heights, this
solution does not generalize to other systems where more spatial dimensions do not exist (e.g. ground robots), and would
not scale well to increasing number of robots or physically constrained environments. 

We approximated the 2D motion of the quadrotors using 2D double integrator dynamics, and thus, $A,B,C$ are given by
\begingroup
    \makeatletter\def\f@size{9}\check@mathfonts
\begin{align}
    A= \left[ \begin{array}{cc}
        I_2 & T_s I_2 \\
        {0_{2,2}} & I_2 \\
    \end{array}\right],\ 
    B= \left[ \begin{array}{c}
            \frac{T_s^2}{2} I_2 \\
         T_s I_2 \\
    \end{array}\right], \ 
    C=\left[ \begin{array}{cc}
        I_2 & {0_{2,2}}
    \end{array}\right], \label{eq:double_integ}
\end{align} 
\endgroup
with sampling time $ T_s=0.1$.  We model the quadrotors as circles ($\agent$ is a circle of radius $r_A=0.1$) to
include the $0.092$ m Crazyflie diameter as well as leave extra margin for aerodynamic effects and a safety padding. 

\textbf{Hardware setup:}
We used six quadrotors ($\numAgents=6$) in our experiments.
We relied on the Crazyswarm platform
\cite{preiss2017crazyswarm} to communicate and control the
quadrotors at 10Hz.  The drones are equipped with
IR-reflective markers detected by an OptiTrack motion
capture system running at 120Hz. The Crazyswarm package
tracked the Crazyflies using the raw point-cloud
data from the OptiTrack motion
capture system, and it issued desired waypoints at a nominal
$10$ Hz update frequency over radio. The Crazyflies tracked those waypoints
using their standard on-board controllers.  In addition to
the uncertainty in the Crazyflie position estimate induced
by the Crazyswarm tracking algorithm, we added a position
estimation noise $\rv{\eta}$ defined in
\eqref{eq:measurement_model}. Such measurement
noises affects the safety filter, but is not
visualized in the plotted physical experiment trajectories.

\textbf{Workspace:}
We considered a $3\times 3$ meter workspace with seven circular obstacles and two goal regions. The obstacles are
depicted by black circles and the goal regions are depicted by transparent circles with a star at the center (see
Figure~\ref{fig:rl_vs_baseline_traj}). We also added position estimation noise to the nominal obstacle locations.

\textbf{Safety filter parameters:} We used Gaussian noise with the following covariances: $\Sigma_{w} =\Sigma_{\eta} =
\mathrm{diag}(10^{-4}, 0, 10^{-4}, 0)$ and $\Sigma_{c_j} = \mathrm{diag}(10^{-4}, 10^{-4}) \ \forall j\in
\Nint{1}{\numObst}$.  As for the risk bounds, we used $\kappa_i = \alpha_{i,j} = \beta_{i,i'} = 0.01$ and divided them
equally across the planning horizon $\horizon=10$. We used $\delta=0.1$ for the terminal 
constraints. For the purposes of constructing the terminal sets, we select velocity bounds of $1$ m/s in the
simulation, and $0.2$ m/s in the experiments. 

\textbf{Computer setup:}
We used an Ubuntu 20.04 LTS workstation with an AMD Ryzen 9 9590X 16-core CPU, a Nvidia GeForce GTX TITAN Black GPU, and
128GB of RAM for all training, simulation, and hardware experiments.

\textbf{RL training:}
We used \texttt{Stable-Baselines3}'s implementation of the PPO (proximal policy optimization) algorithm~\cite{raffin2019stable} to train the RL agents. 
We ran two training sessions, one for each goal, for $10$ million time steps each. We used the default parameters of \texttt{Stable-Baselines3} with the following modifications: $0.01$ entropy coefficient, $2021$ seed, and \texttt{cpu} device. 
We use $\zeta_\text{obs} = -0.001, \ \zeta_\text{tgt} = -0.1,\ R_\text{target} = 10^4, \ P_\text{keep-in} = -10^4$, and $P_\text{obstacle} = -500$ for the reward function parameters.

After training, we selected the trained policy at about $9.7$ 
million steps and $9.44$ million steps for the two targets
respectively. Each training session took just over $11$ hours. 

\textbf{Choice of unit vectors in Proposition~\ref{prop:risk_tightened_constraints},~\ref{prop:risk_tightened_terminal_constraints}:}
Inspired by~\cite{augugliaro2012generation}, we used the following unit vectors:
\begingroup
    \makeatletter\def\f@size{8}\check@mathfonts
    \begin{subequations}
    \begin{align}
        \zijobs(k|t) &\triangleq \frac{p^{\text{RL}}_i(k|t) - \mean{c}_j}{\|p^{\text{RL}}_i(k|t) - \mean{c}_j\|},
        \zijagent(k|t) \triangleq \frac{p^{\text{RL}}_i(k|t) - p^{\text{RL}}_j(k|t)}{\|p^{\text{RL}}_i(k|t) - p^{\text{RL}}_j(k|t)\|} \\
        \ellijobs(k|t) &\triangleq \frac{x^{\text{RL}}_i(k|t) - \mean{c}_j^{\text{lift}}}{\|x^{\text{RL}}_i(k|t) - \mean{c}_j^{\text{lift}}\|}, 
        \ellijagent(k|t) \triangleq \frac{x^{\text{RL}}_i(k|t) - x^{\text{RL}}_j(k|t)}{\|x^{\text{RL}}_i(k|t) - x^{\text{RL}}_j(k|t)\|}
    \end{align}\label{eq:convexification_vec}%
    \end{subequations}%
\endgroup
where $p^{\text{RL}}_i(k|t) = C x^{\text{RL}}_i(k|t)$. Such
a choice used the predicted RL states and positions and the
nominal obstacle locations to produce a heuristic for the
computation of the safe halfspace polytopes. For
the 2D double integrator dynamics~\eqref{eq:double_integ},
the lifted state is the position vector with zeros
appended for the velocity components, i.e.
$\mean{c}_j^{\text{lift}} = [\mean{c}_j^\top \
0_2^\top]^\top$.

\textbf{Solving the QP:} 
We modeled the QP associated with the safety filter in
Python 3.7 using CVXPY~\cite{diamond2016cvxpy}, utilizing
parameters for values in~\eqref{prob:safety_filter_reform}
that change at every control time step, and solved it using
ECOS~\cite{domahidi2013ecos} in experiments,
and~\texttt{GUROBI}~\cite{GUROBI} in simulations.

\section{Experiments} \label{sec:experiments}
We present results of the experimental validation of our approach on a quadrotor testbed. We show that the
trained, single-agent RL-based motion planner generalizes well when used with the proposed safety filter.  We also
compare the proposed approach with a MPC-based multi-agent motion planner in simulation to emphasize the benefits of the
RL step as well as the effects of the terminal constraints. We conclude with a demonstration of the scalability of our
approach.

\subsection{Experimental validation}

Figure~\ref{fig:rl_vs_baseline_experiment} shows snapshots of two experiments and their reconstructed plots.  In these
experiments, we compare the proposed solution with a safety-filtered baseline controller. Here, the baseline controller
is a proportional controller that regulates the drones to the target while ignoring all static and dynamic obstacles,
which are handled by the safety filter \eqref{prob:safety_filter_reform}.

In Figure~\ref{fig:rl_vs_baseline_experiment}, the top two rows are for the proposed solution with the RL controller and
the proposed safety filter \eqref{prob:safety_filter_reform} while the bottom two rows use the baseline controller
instead of the RL controller.  In both cases, the proposed safety filter ensures that the agents remain safe.  In the RL
case, the agents manage to reach their goals more rapidly, while the baseline controller case, the agents take
significantly longer to reach their goals. In fact, when using the baseline controller instead of the RL controller, we
found that the pink agent typically gets stuck between two obstacles and fails to reach its goal  (see the bottom two
rows of Figure \ref{fig:rl_vs_baseline_experiment}). 

\begin{figure}[t]
    \centering
    \includegraphics[trim={10cm 0cm 8cm 0cm},clip, width=0.32\linewidth]{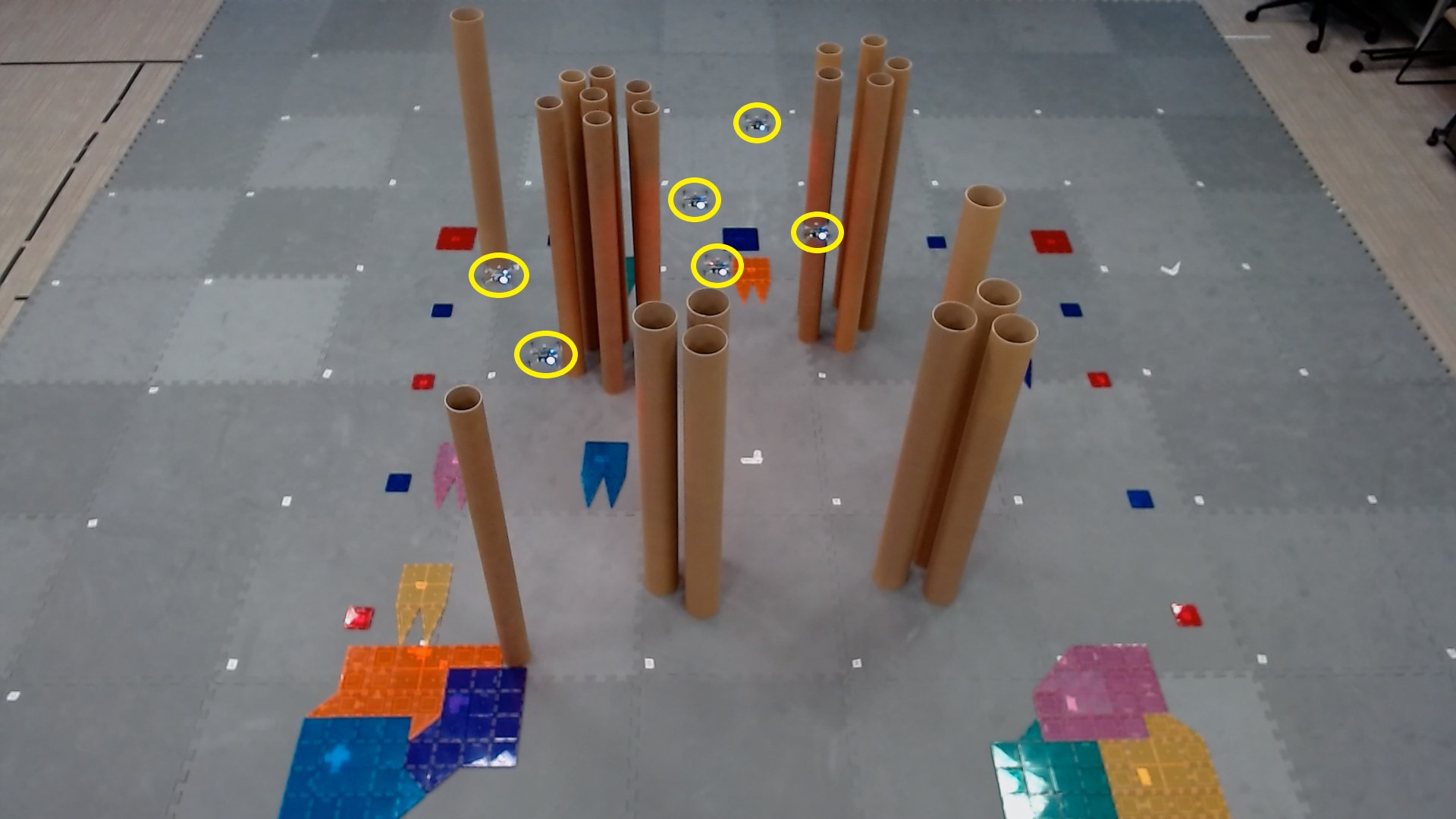}
    \includegraphics[trim={10cm 0cm 8cm 0cm},clip, width=0.32\linewidth]{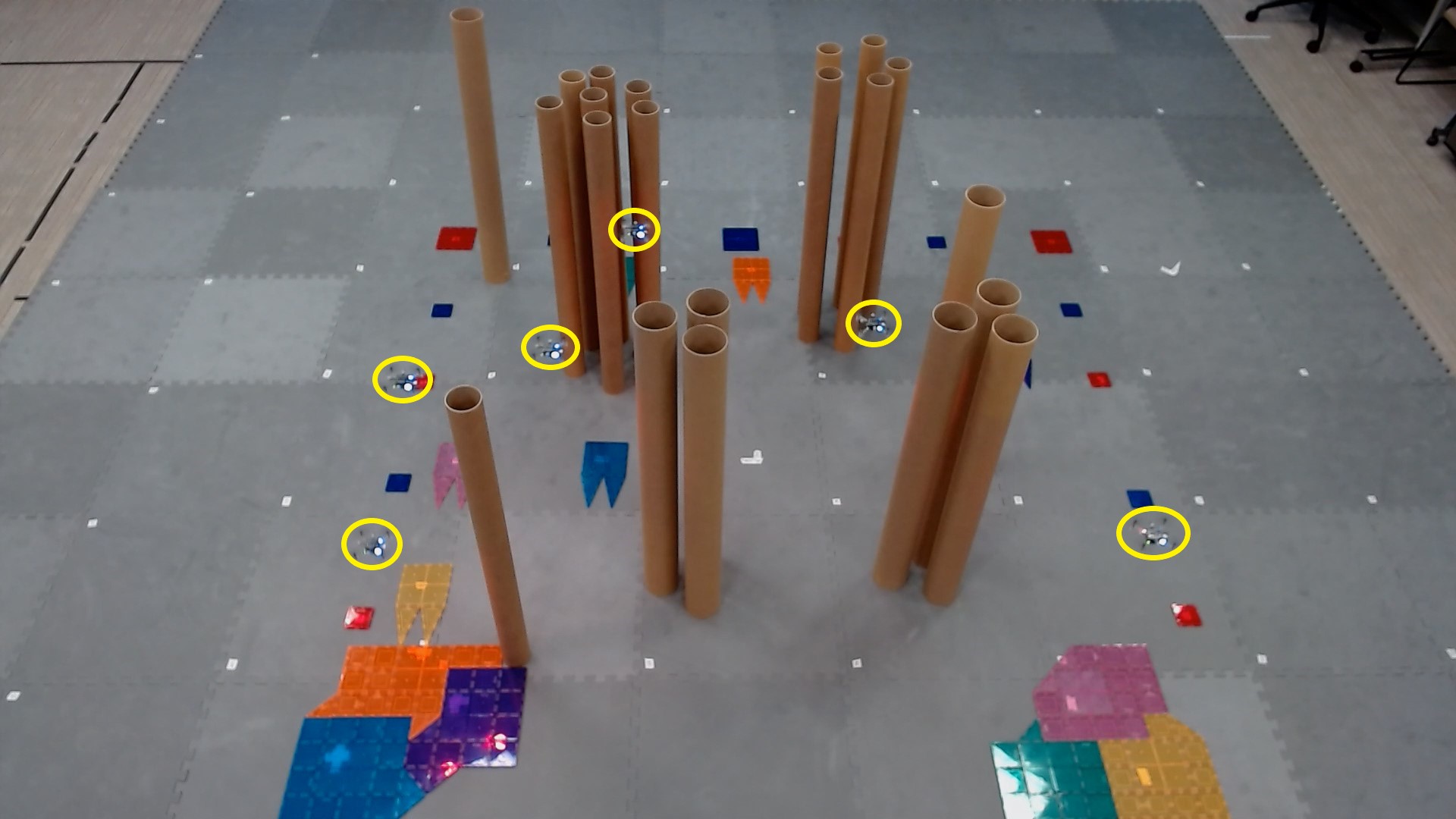}
    \includegraphics[trim={10cm 0cm 8cm 0cm},clip, width=0.32\linewidth]{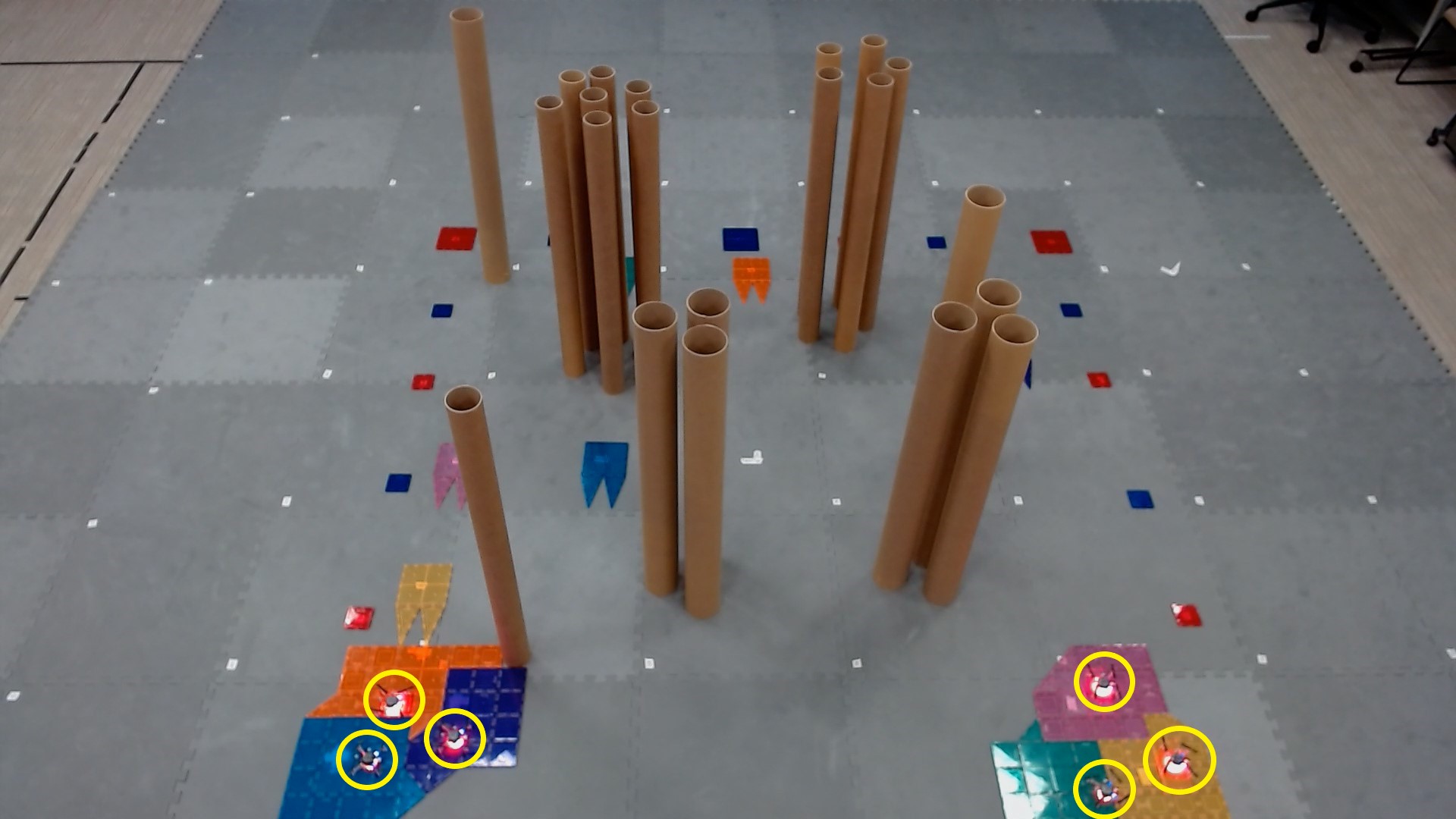} \\
    \includegraphics[trim={13cm 5cm 13cm 5cm},clip, width=0.32\linewidth]{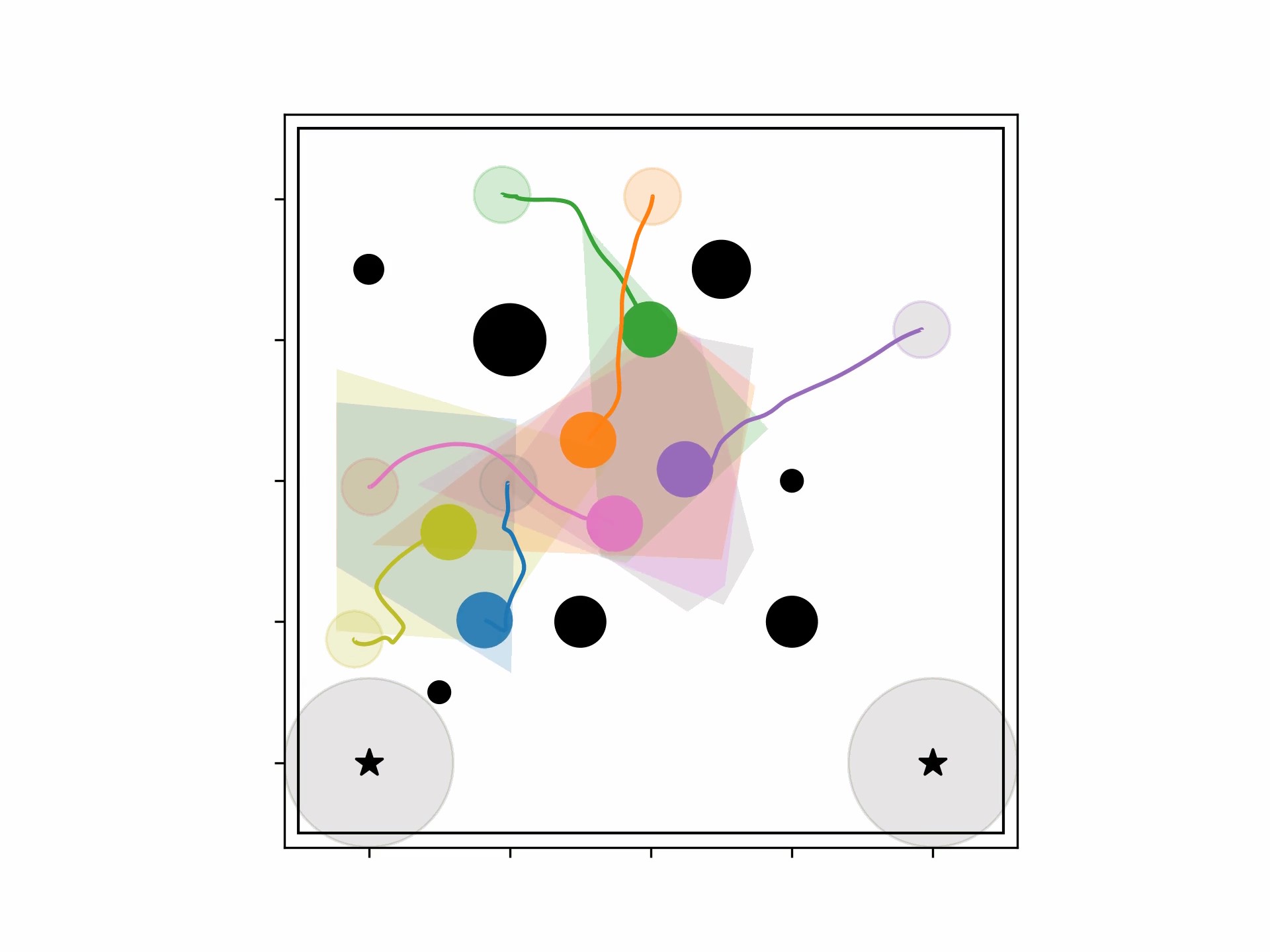}
    \includegraphics[trim={13cm 5cm 13cm 5cm},clip, width=0.32\linewidth]{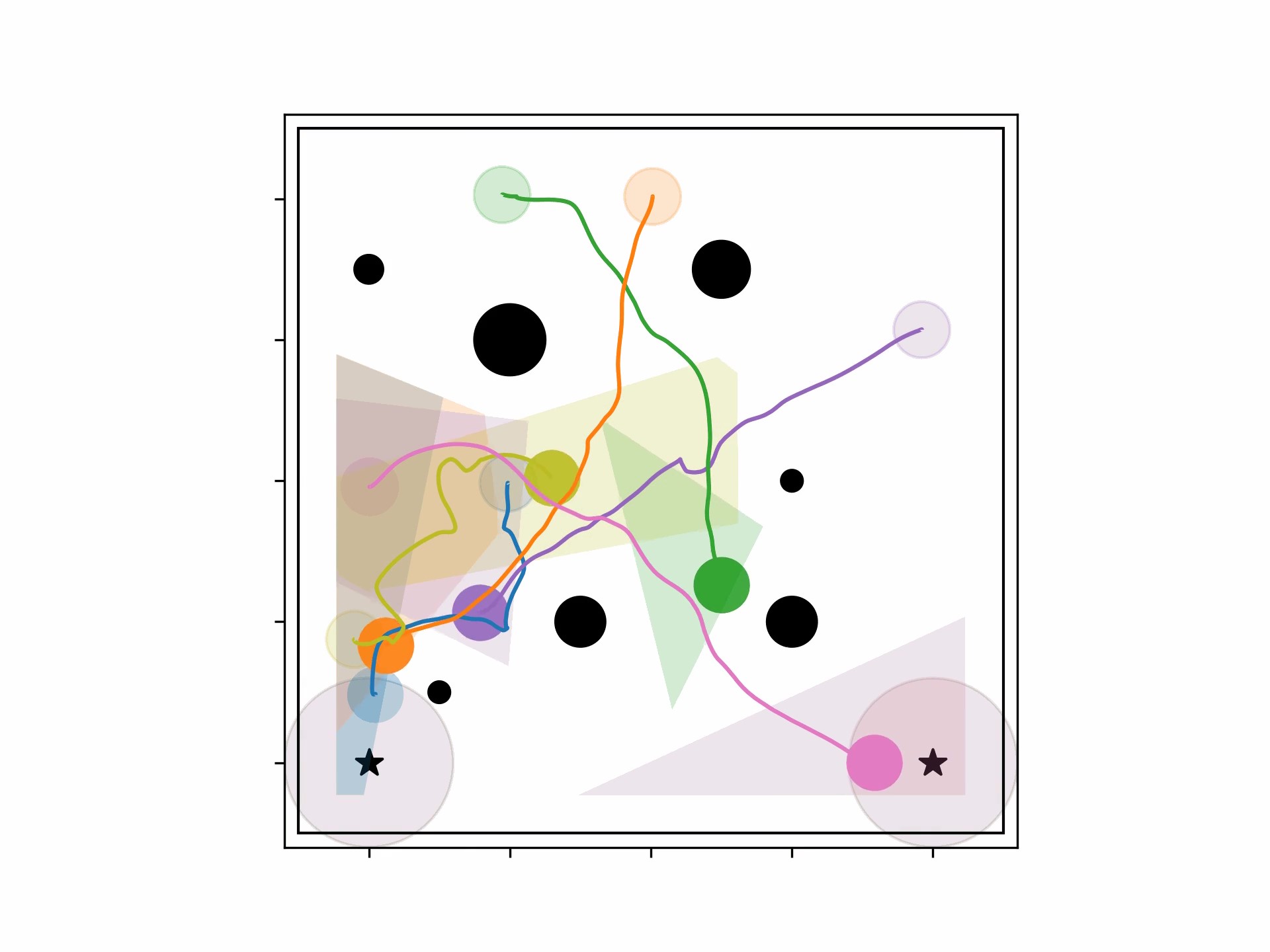}
    \includegraphics[trim={13cm 5cm 13cm 5cm},clip, width=0.32\linewidth]{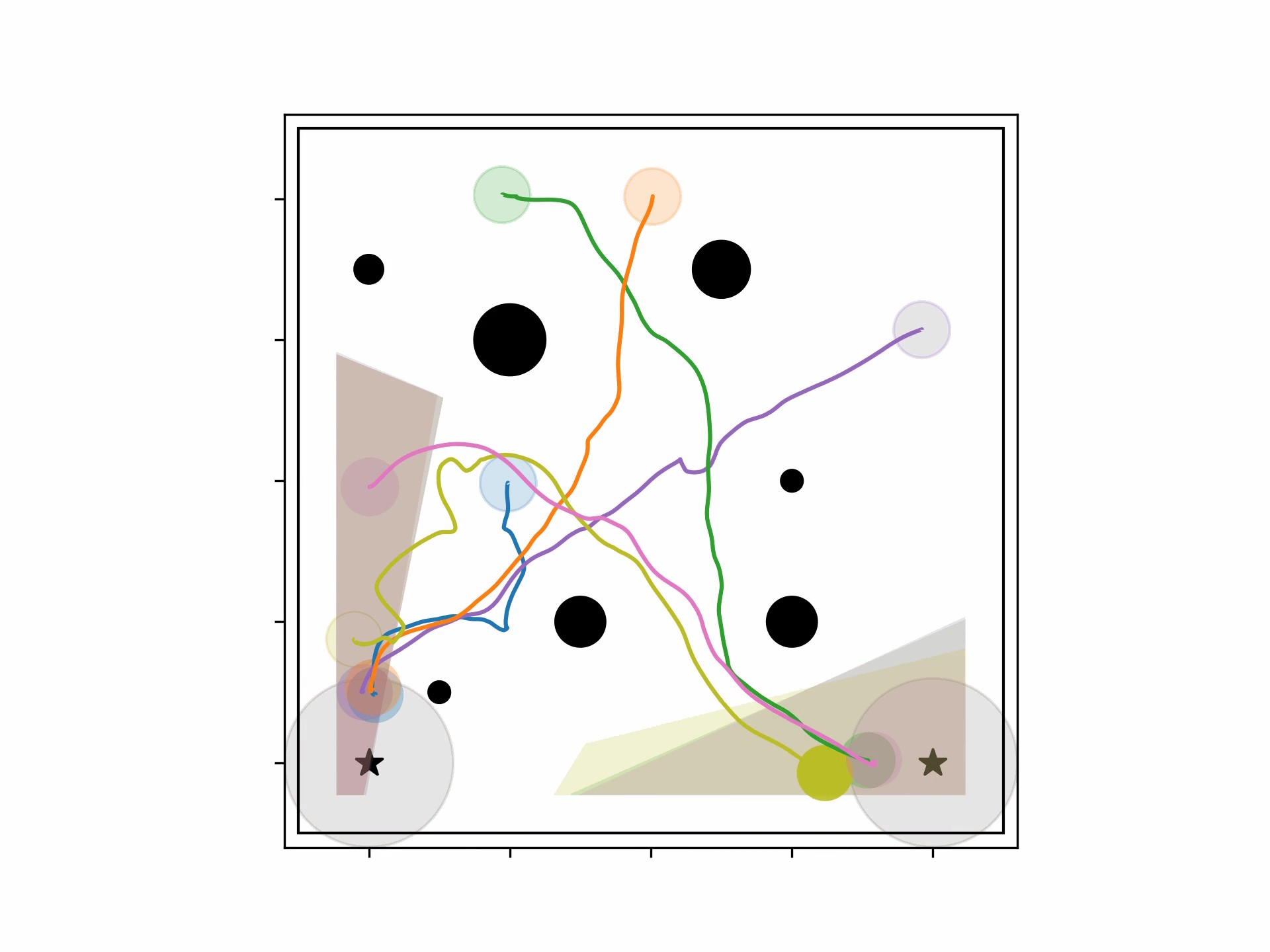} \\
    \includegraphics[trim={13cm 5cm 13cm 5cm},clip, width=0.32\linewidth]{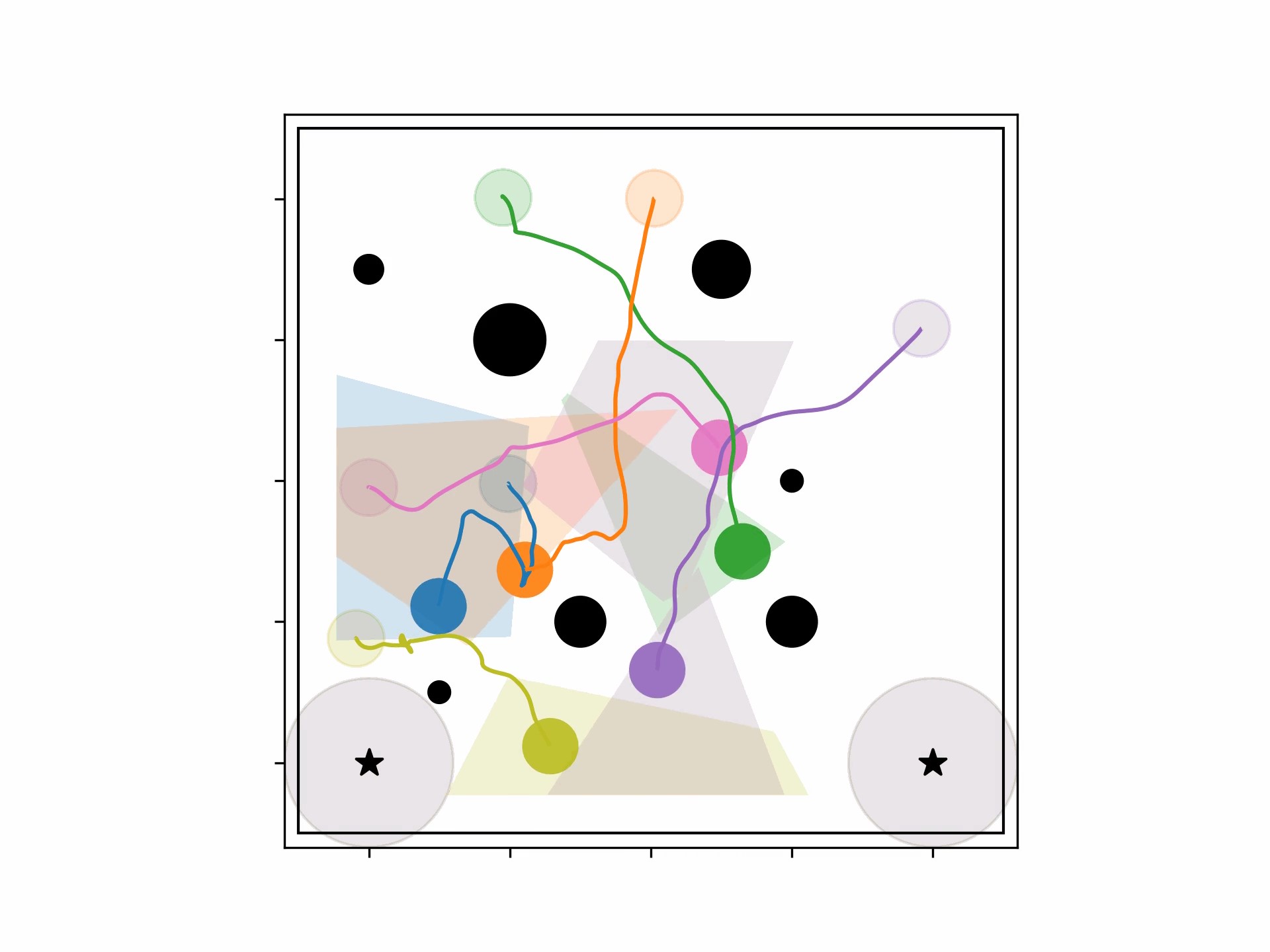}
    \includegraphics[trim={13cm 5cm 13cm 5cm},clip, width=0.32\linewidth]{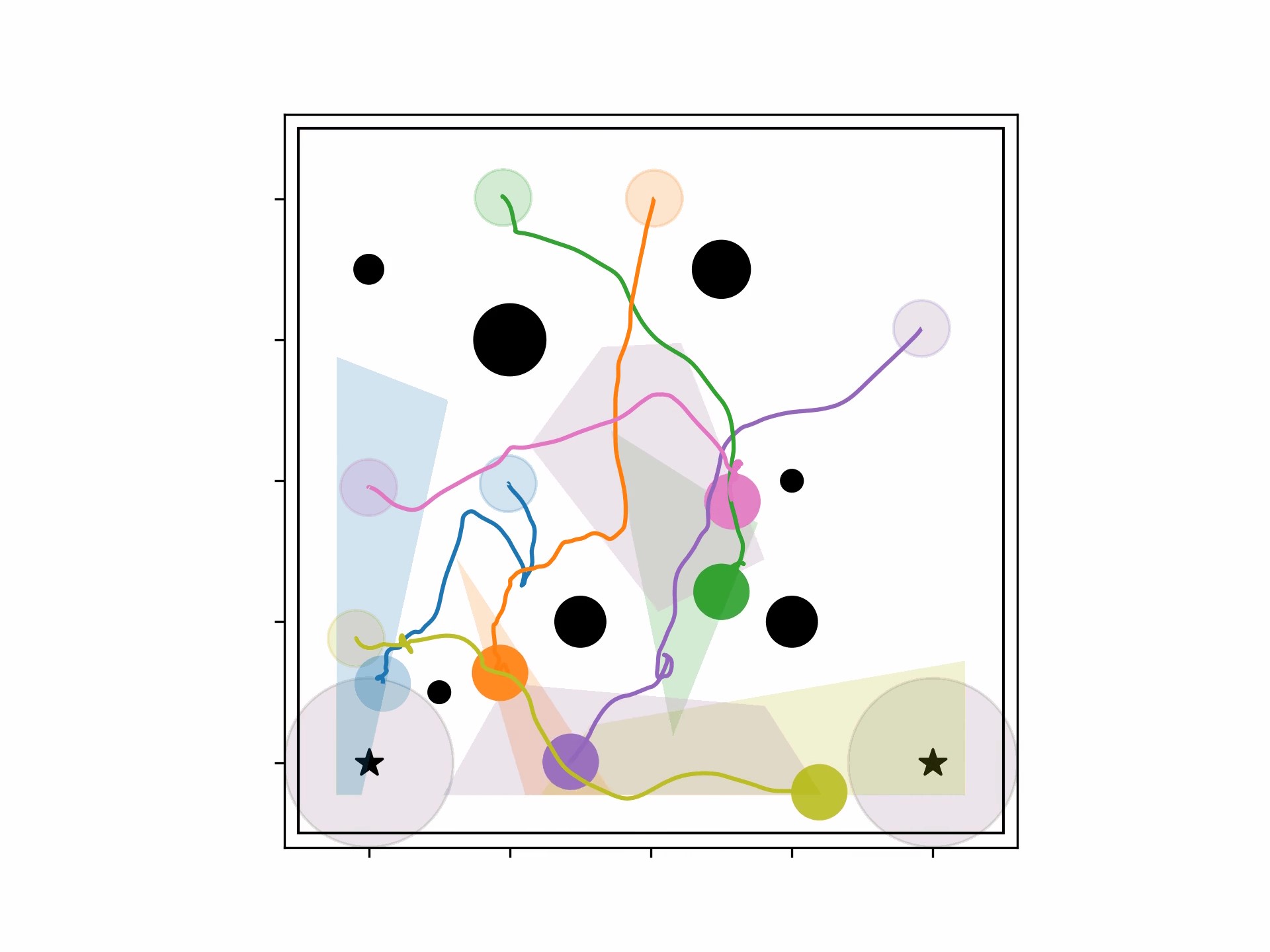}
    \includegraphics[trim={13cm 5cm 13cm 5cm},clip, width=0.32\linewidth]{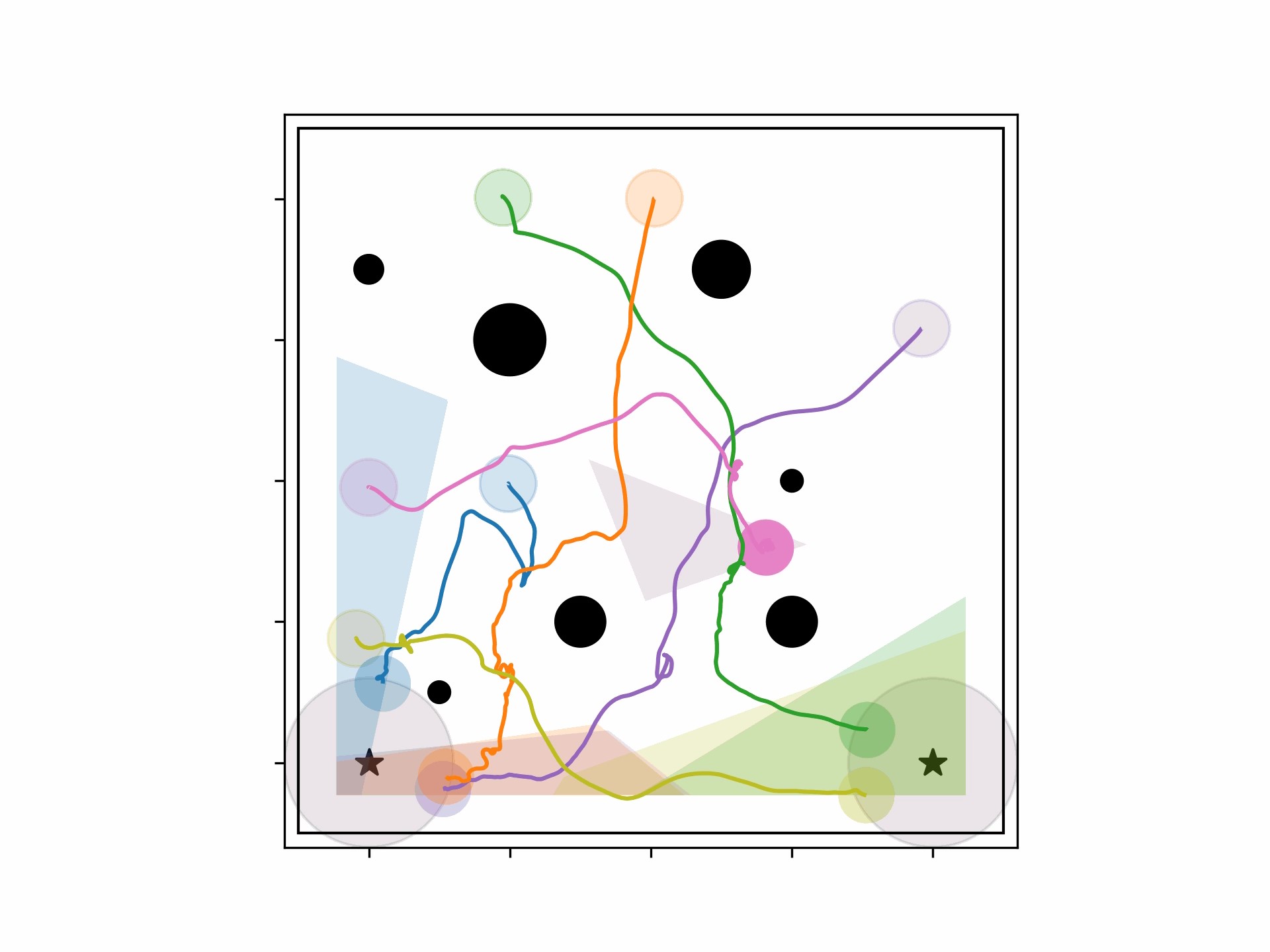} \\
    \includegraphics[trim={10cm 0cm 8cm 0cm},clip, width=0.32\linewidth]{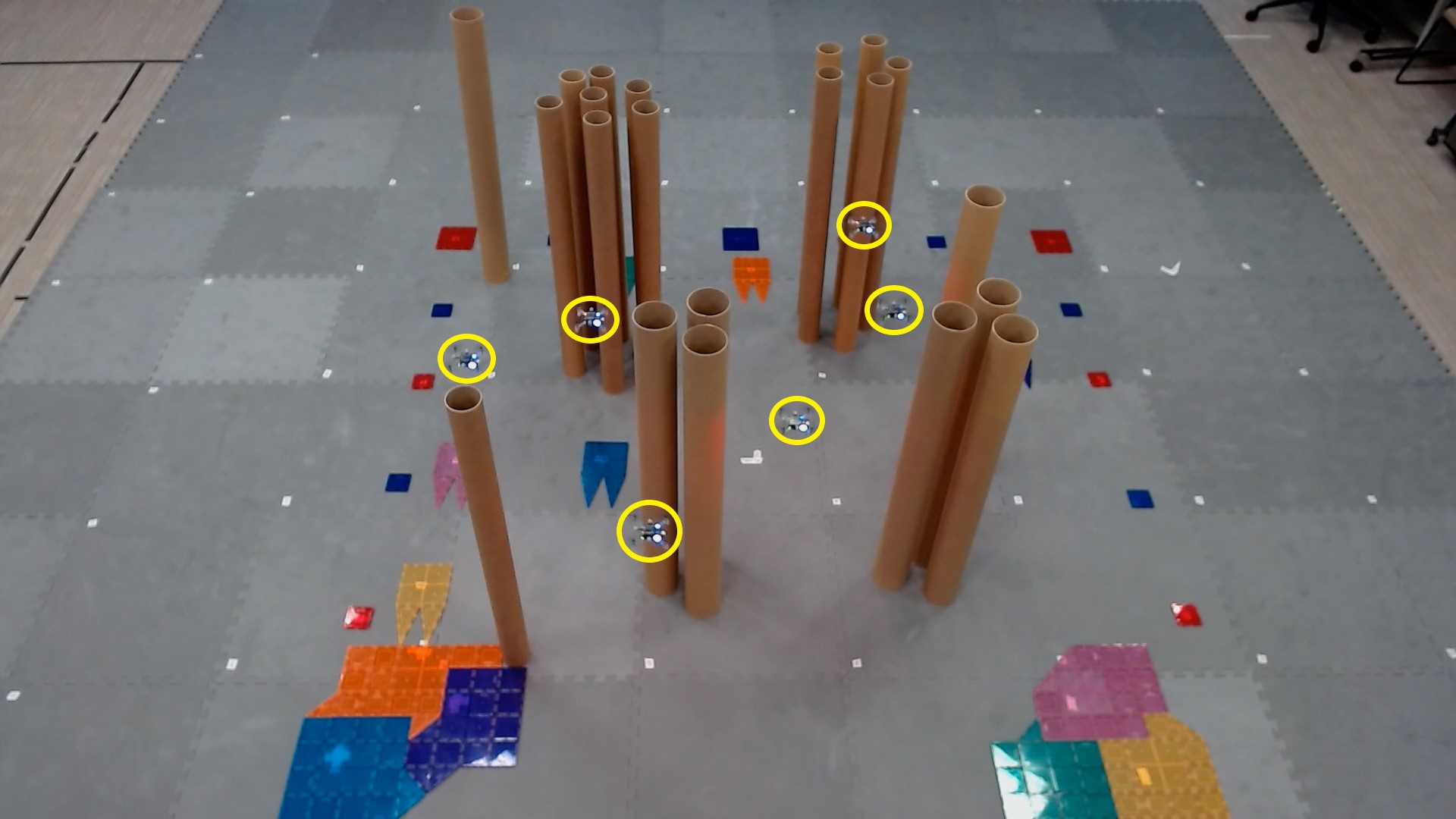}
    \includegraphics[trim={10cm 0cm 8cm 0cm},clip, width=0.32\linewidth]{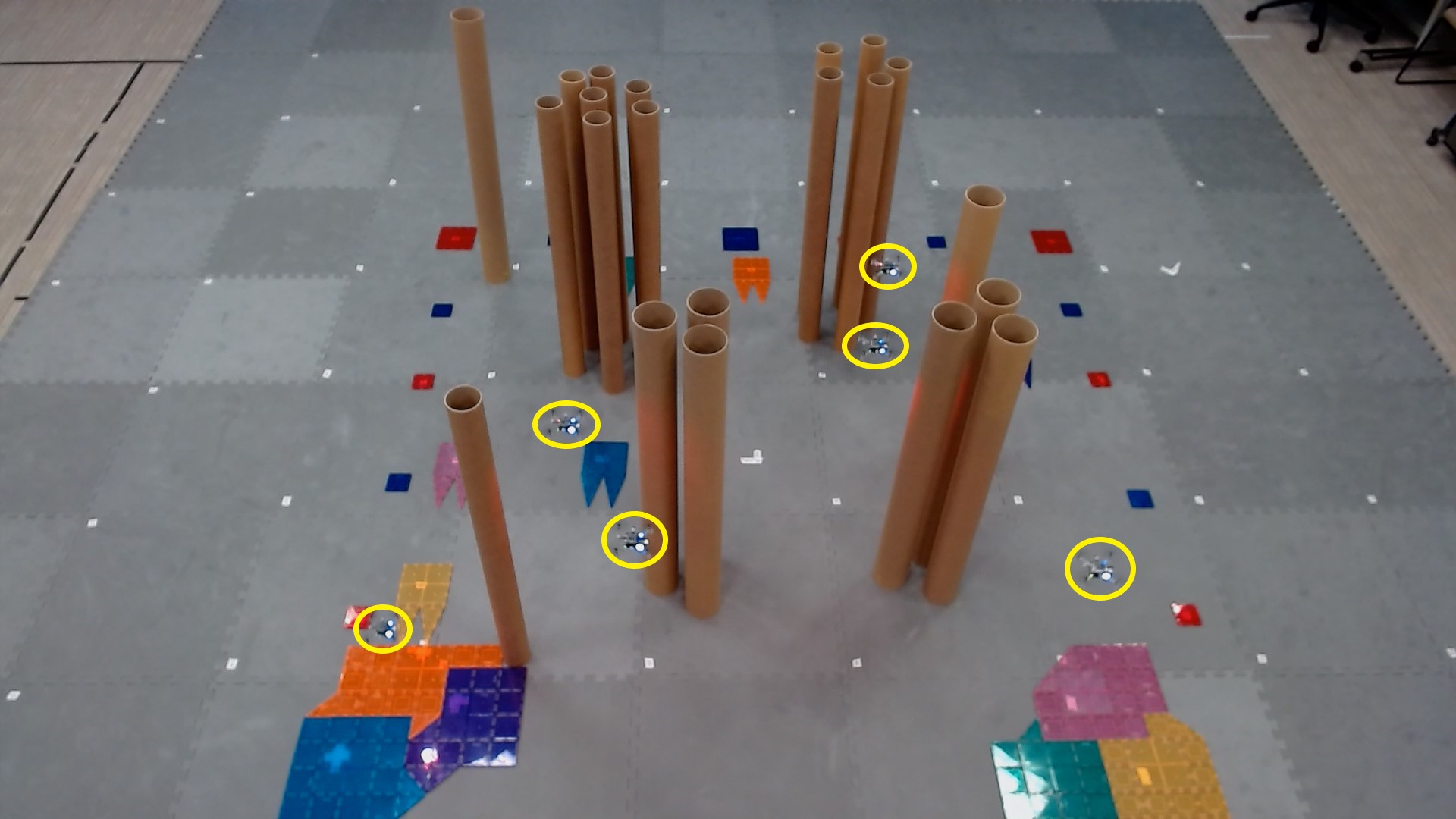}
    \includegraphics[trim={10cm 0cm 8cm 0cm},clip, width=0.32\linewidth]{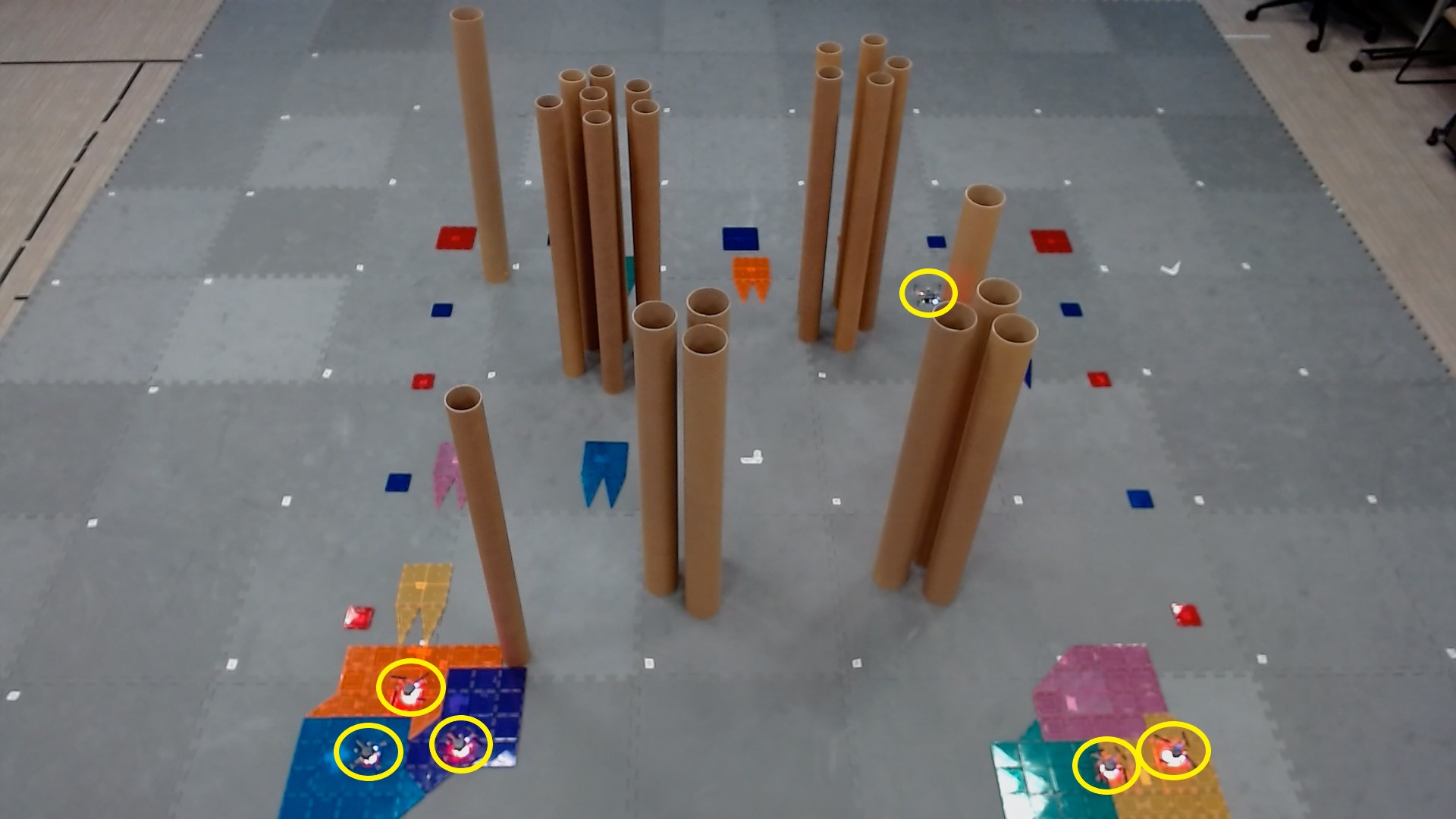} \\
    \caption{Safe multi-agent motion planning using the
        proposed safety filter in conjunction with the
        RL-based controller and a classical proportional
        controller (baseline).  (Top two rows) Snapshots and
        reconstructed illustrations of the hardware
        experiment's trajectories when using the RL-based
        controller with the safety filter at $7$, $14$, and
        $21$ seconds.  (Bottom two rows) Trajectories of the
        hardware experiment when using the baseline
        controller with the safety filter at times $13$,
        $21$, and $70$ seconds.  The black circles and
        boundary are the obstacles and keep-in set.
        Transparent starred circles depict the targets.
        Colored circles denote the agents' starting and goal
        positions. 
    The colored paths indicate the trajectory and the shaded
regions are the static obstacle-free positions at the
current control time step (determined via
convexification).}\label{fig:rl_vs_baseline_experiment}
\end{figure}

\begin{figure}[htbp]
    \centering
    \includegraphics[trim={0.7cm 0cm 1.3cm 1.35cm},clip,
    width=1\linewidth]{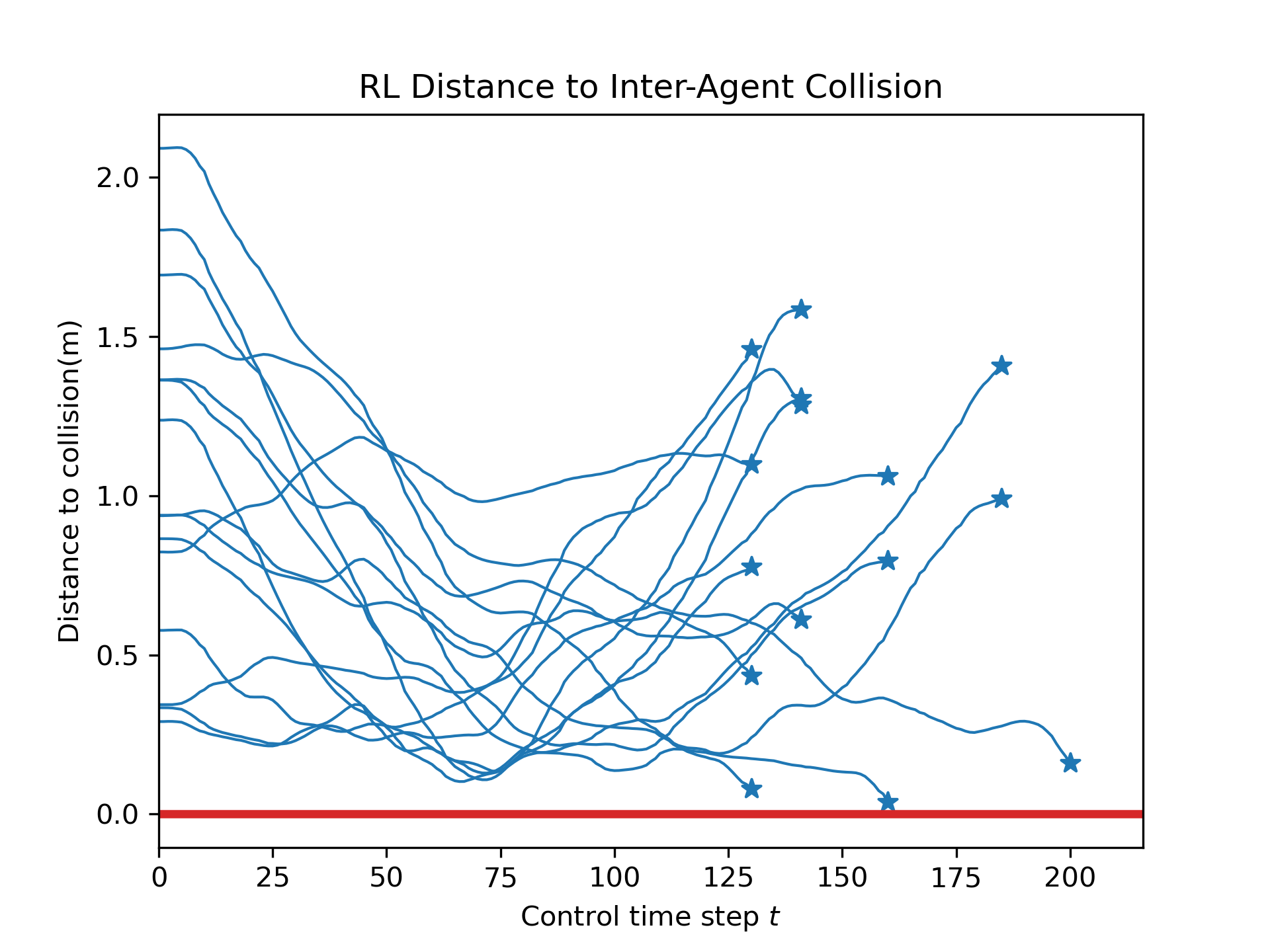}
    \caption{Clearance between the agents during the
        physical experiment with RL controller, where a
        clearance (distance to collision) accounts for the
        physical dimensions of the agents. A negative
    clearance indicates a collision.  Stars indicate one of
the two agents reaching the target.}
\label{fig:inter_agent_dists}
\end{figure}

Figure~\ref{fig:inter_agent_dists} shows the clearances
between each agent ($15$ pairs for the six agents) during
the physical experiment.  Specifically, it plots the
inter-agent distances \emph{minus} twice the agent radius,
i.e. $ \|p_i(t) - p_j(t) \| - 2\agentRad \ \forall i,j, i
\neq j$. Thus, a negative distance indicates a collision.
Due to the use of probabilistic constraints, the distances
are always positive, which shows that the system is
collectively safe.

Figure~\ref{fig:timing_graph} shows the QP setup time (blue) and the total time for setting up and solving the QP (orange) over the RL experiment's duration.
The total time spent setting up and solving \eqref{prob:safety_filter_reform} for six agents was on average $0.05$
seconds. Since the time spent was always less than $0.06$ seconds, we had a sufficient margin to the $0.1$ control sampling period.

\begin{figure}
    \centering
    \includegraphics[trim={0.9cm 1.95cm 0.8cm 3.05cm},clip,
    width=0.9\linewidth]{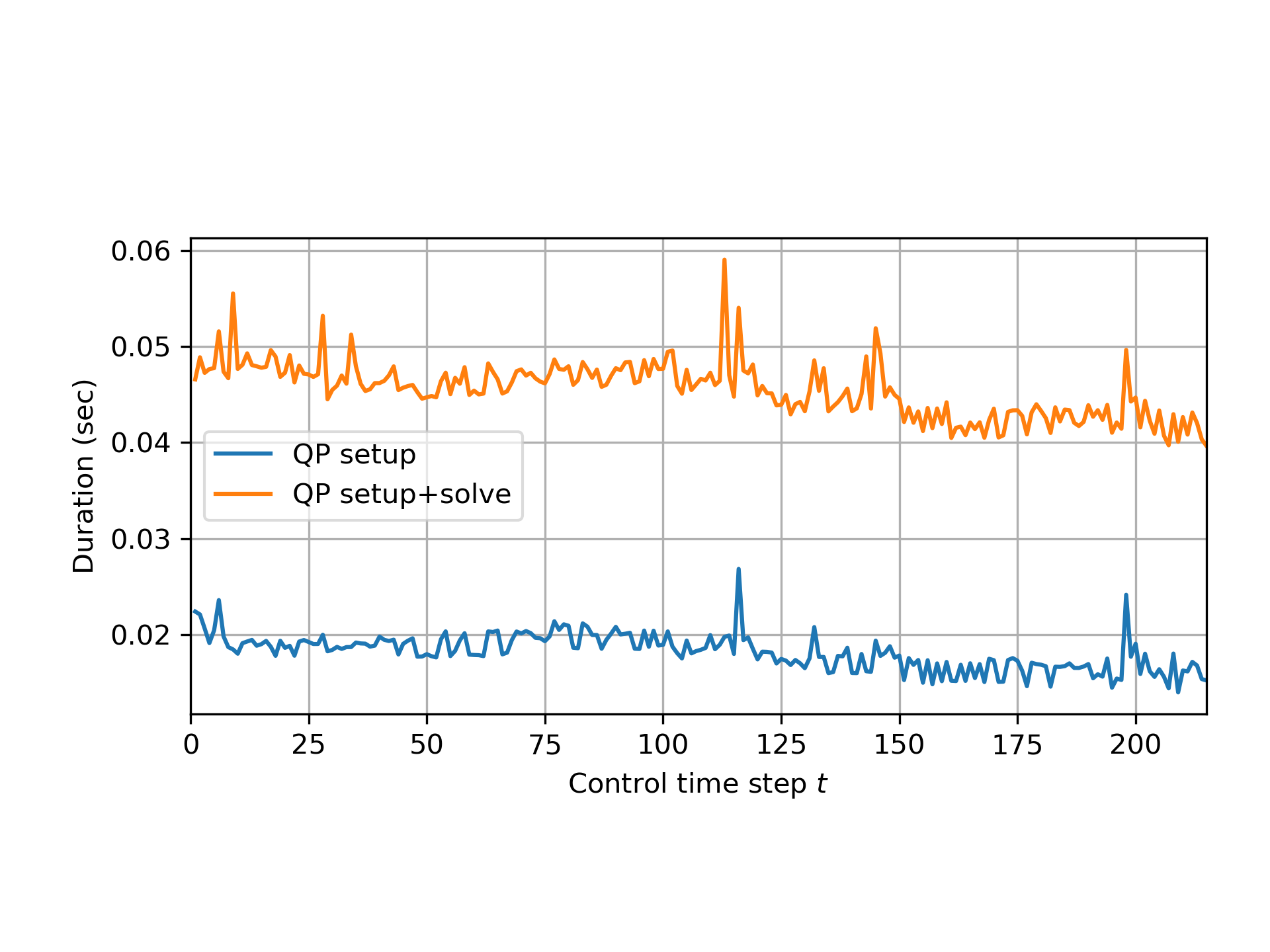}
    \caption{Problem setup and solution durations  to solve
    the quadratic program
\eqref{prob:safety_filter_reform} in the experiment
using CVXPY~\cite{diamond2016cvxpy} and
ECOS~\cite{domahidi2013ecos}.}\label{fig:timing_graph}
\end{figure}

Figure~\ref{fig:rl_vs_baseline_traj} shows the
reconstruction of the agent trajectories for both the RL and
baseline controllers based on the data collected during the
experiments. As expected, the final trajectories for both RL
and baseline controllers remain sufficiently far from
the obstacles and the keep-in set bounds. While avoiding the
red padding, which represents the enlargement of the
obstacle rigid body by the agent's radius, is sufficient for
collision avoidance, the chance constraints prevent the
trajectories from getting too close and hence result in the
additional virtual padding around the obstacles.

\begin{figure}
    \centering
    \includegraphics[trim={3cm 1cm 3cm 0.5cm},clip, width=0.49\linewidth]{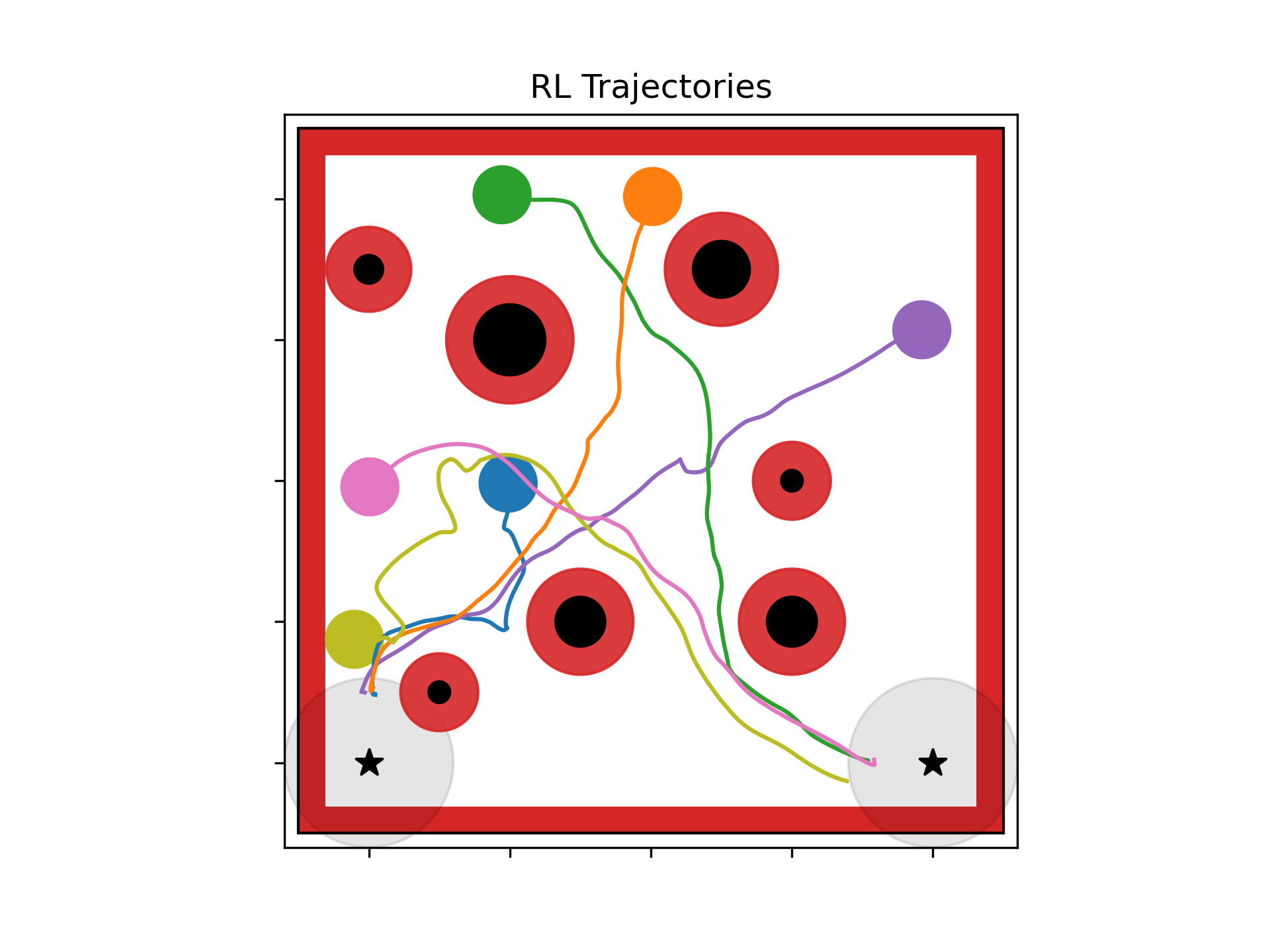}
    \includegraphics[trim={3cm 1cm 3cm 0.5cm},clip, width=0.49\linewidth]{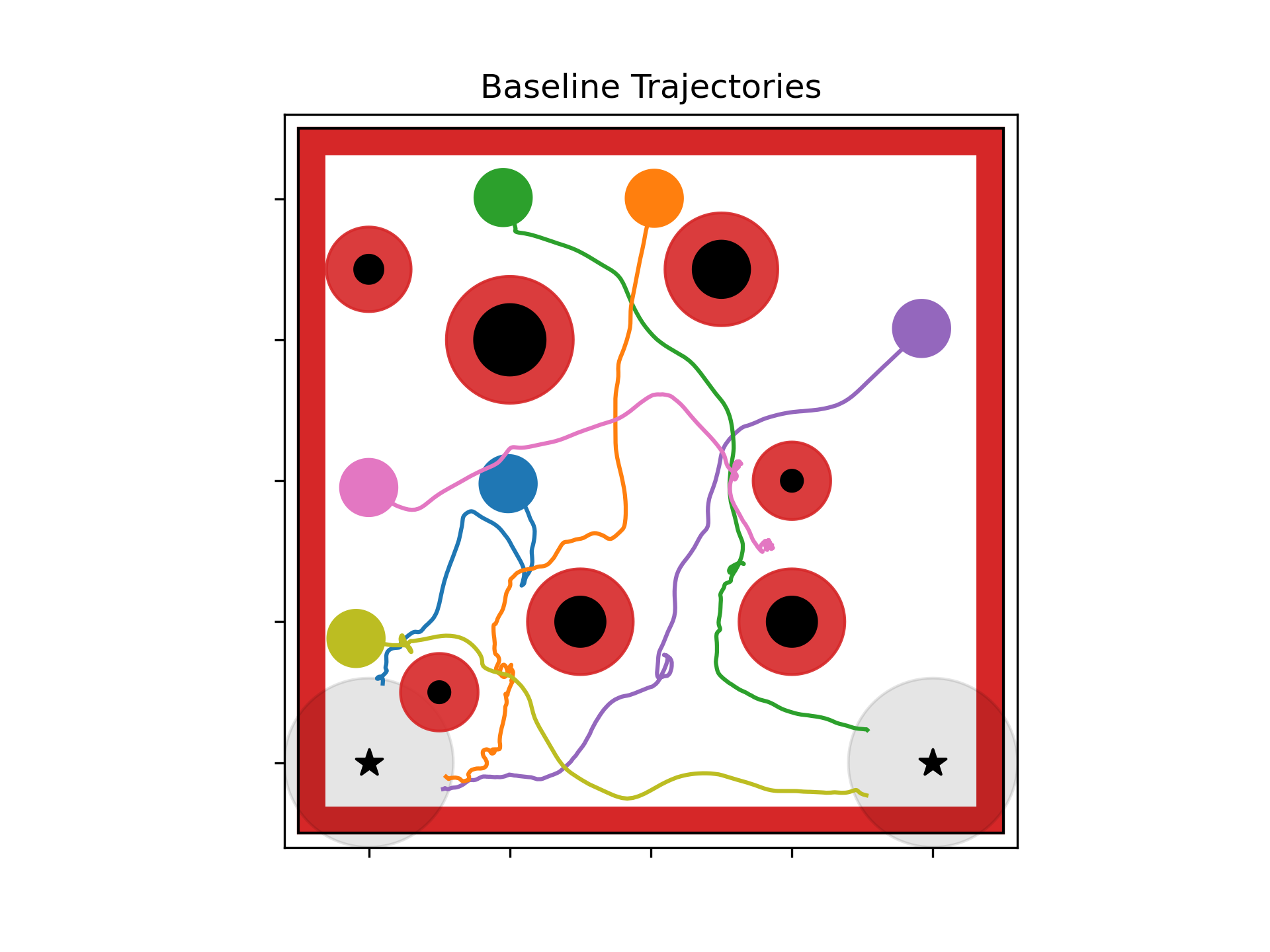}
    \caption{Reconstruction of the RL (left) and baseline
        (right) trajectories from the experiments.
    The red padding around the keep-in set and obstacles, representing the agent radius, is never crossed and hence all trajectories are safe.}
    \label{fig:rl_vs_baseline_traj}
\end{figure}

\subsection{Evaluation of the RL motion planner}

The deterministic evaluation of the learned policy over a
$100\times 100$ grid is presented in
Figure~\ref{fig:rl_eval} (top row). 

We observe that the RL agents learned to navigate to the
goal starting from most initial conditions. As expected, the
learned policy is not perfect and sometimes results in
collisions with the obstacles or the workspace
(Rows~$1$ and~$3$). Nevertheless, the combination of RL and
the safety filter ensures safe motion planning (Rows $2$ and
$4$). For less than $2\%$ of the initial conditions, the RL
policy did not reach the target within $800$ time steps
($80$ s), which we mark as ``loiter", i.e., static/dynamic deadlock, but safety was still
guaranteed. In practice, it is usually possible to recover from such deadlock conditions by small state perturbations. We plan to investigate formal methods for avoiding and recovering from such deadlocks in future studies.

\begin{figure}
    \centering
    \includegraphics[trim={2cm 9cm 2cm 0cm},clip, width=0.9\linewidth]{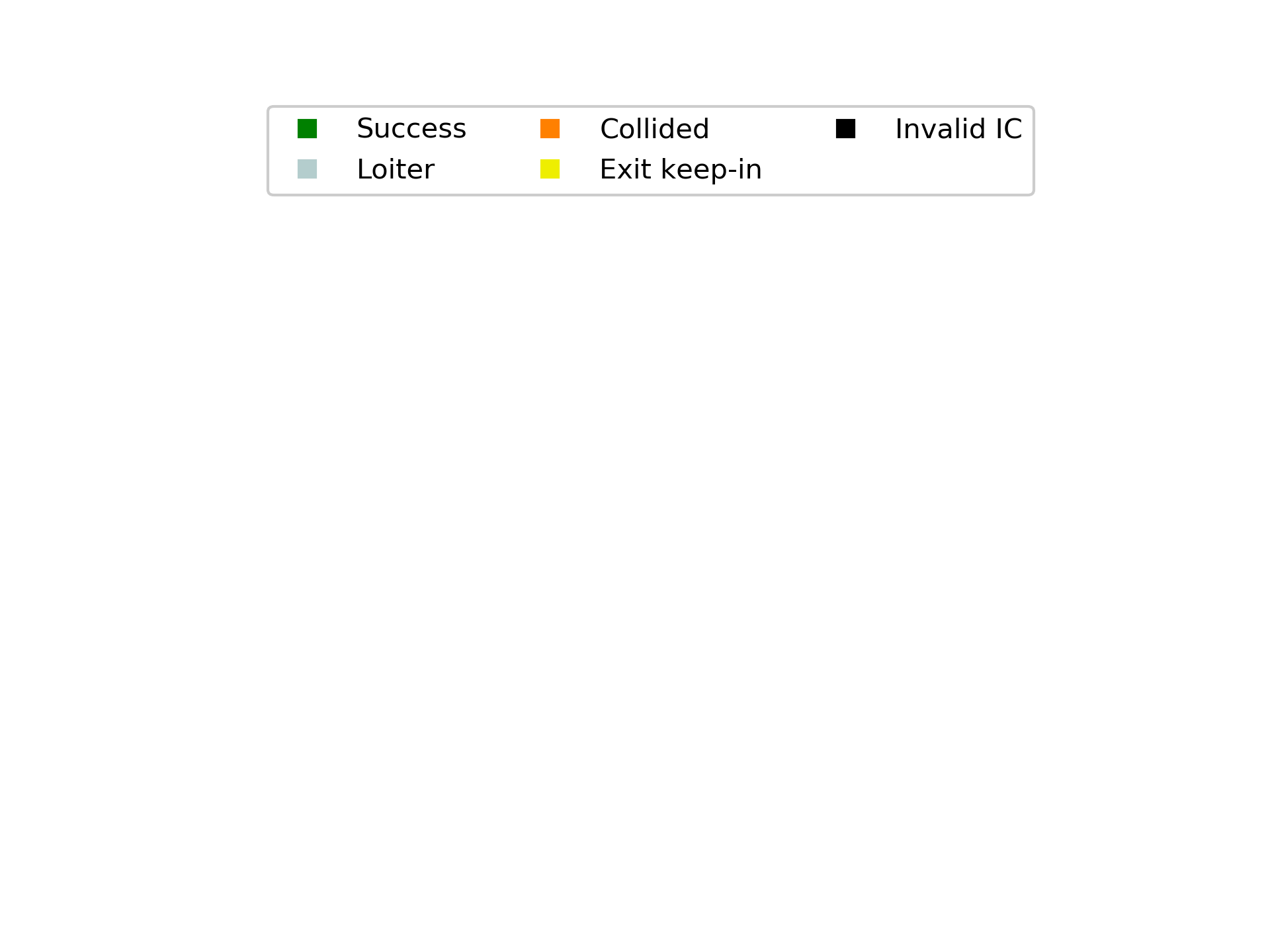} \\
    \includegraphics[trim={3cm 1cm 3cm 0.5cm},clip, width=0.49\linewidth, angle=180,origin=c]{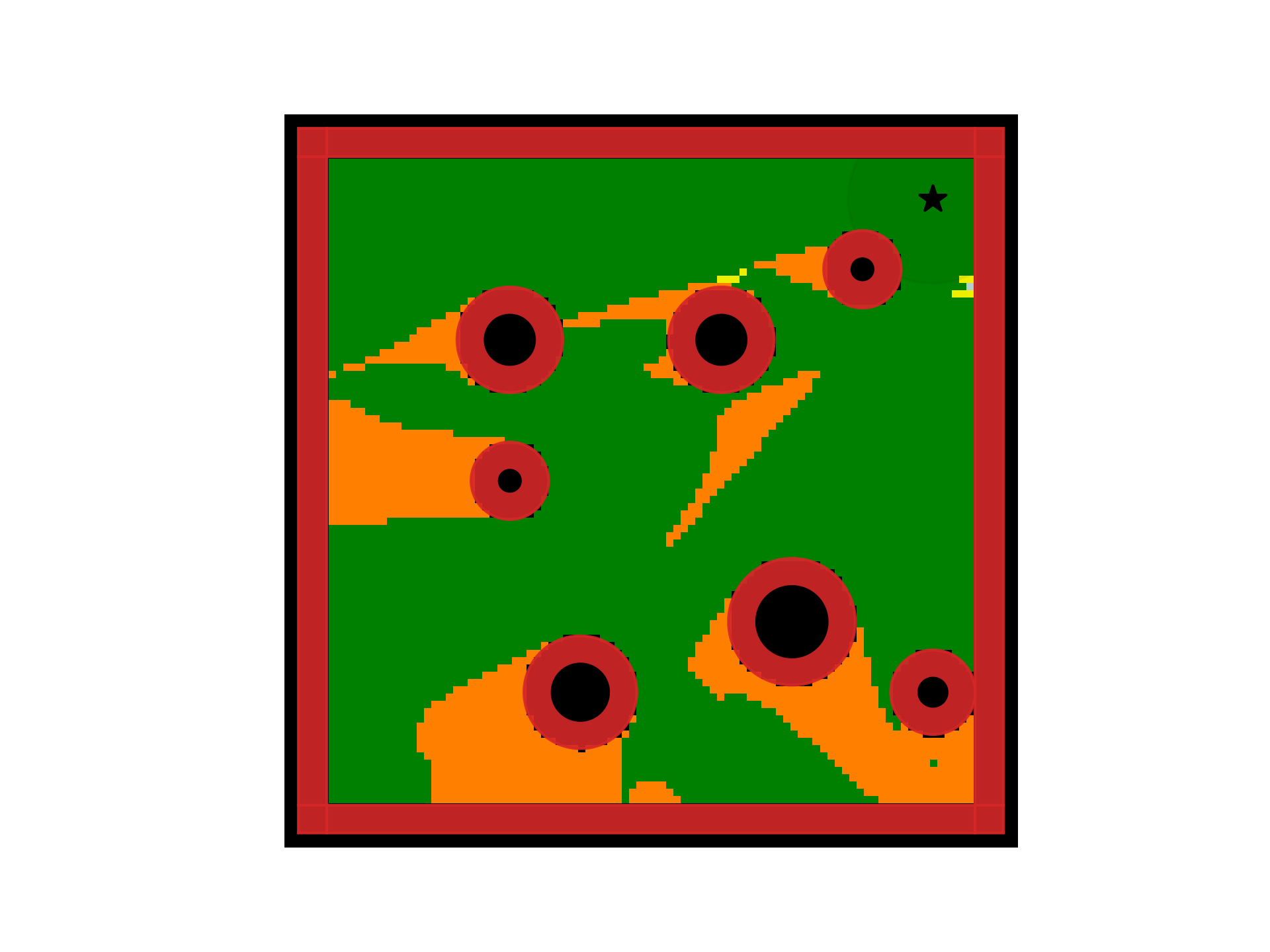}
    \includegraphics[trim={3cm 1cm 3cm 0.5cm},clip, width=0.49\linewidth, angle=180,origin=c]{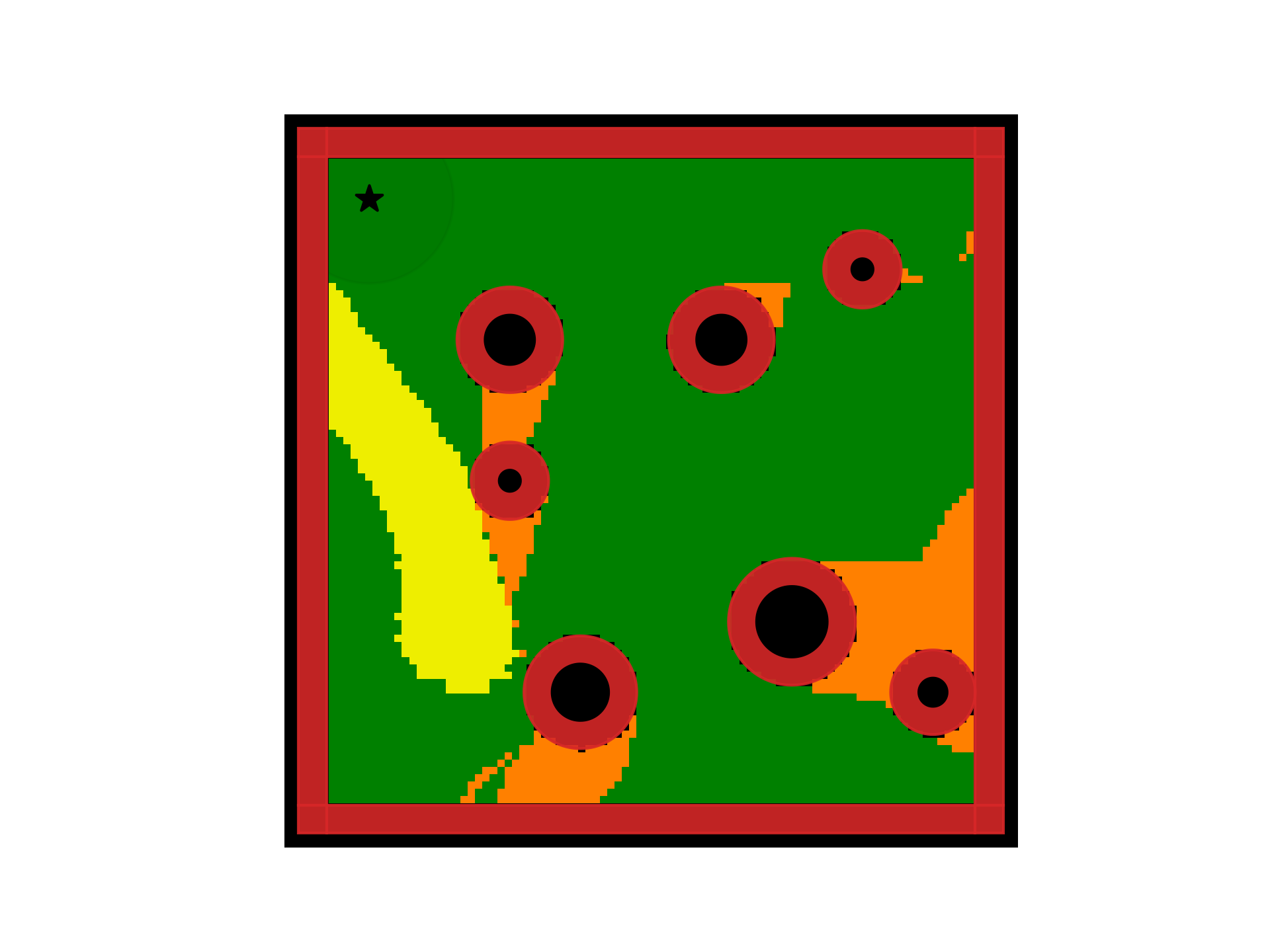} \\
    \includegraphics[trim={3cm 1cm 3cm 0.5cm},clip,width=0.49\linewidth, angle=180,origin=c]{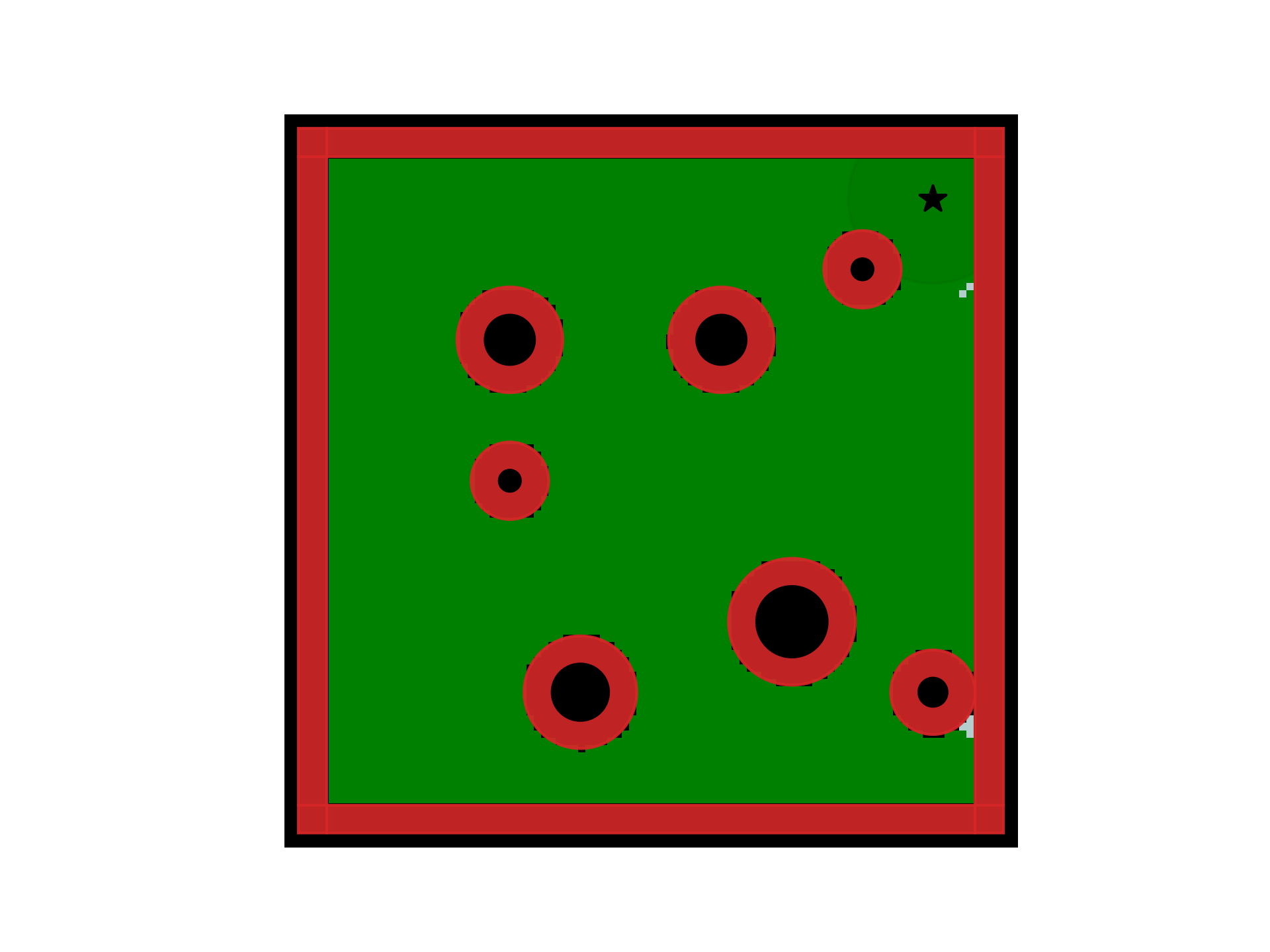}
    \includegraphics[trim={3cm 1cm 3cm 0.5cm},clip,width=0.49\linewidth, angle=180,origin=c]{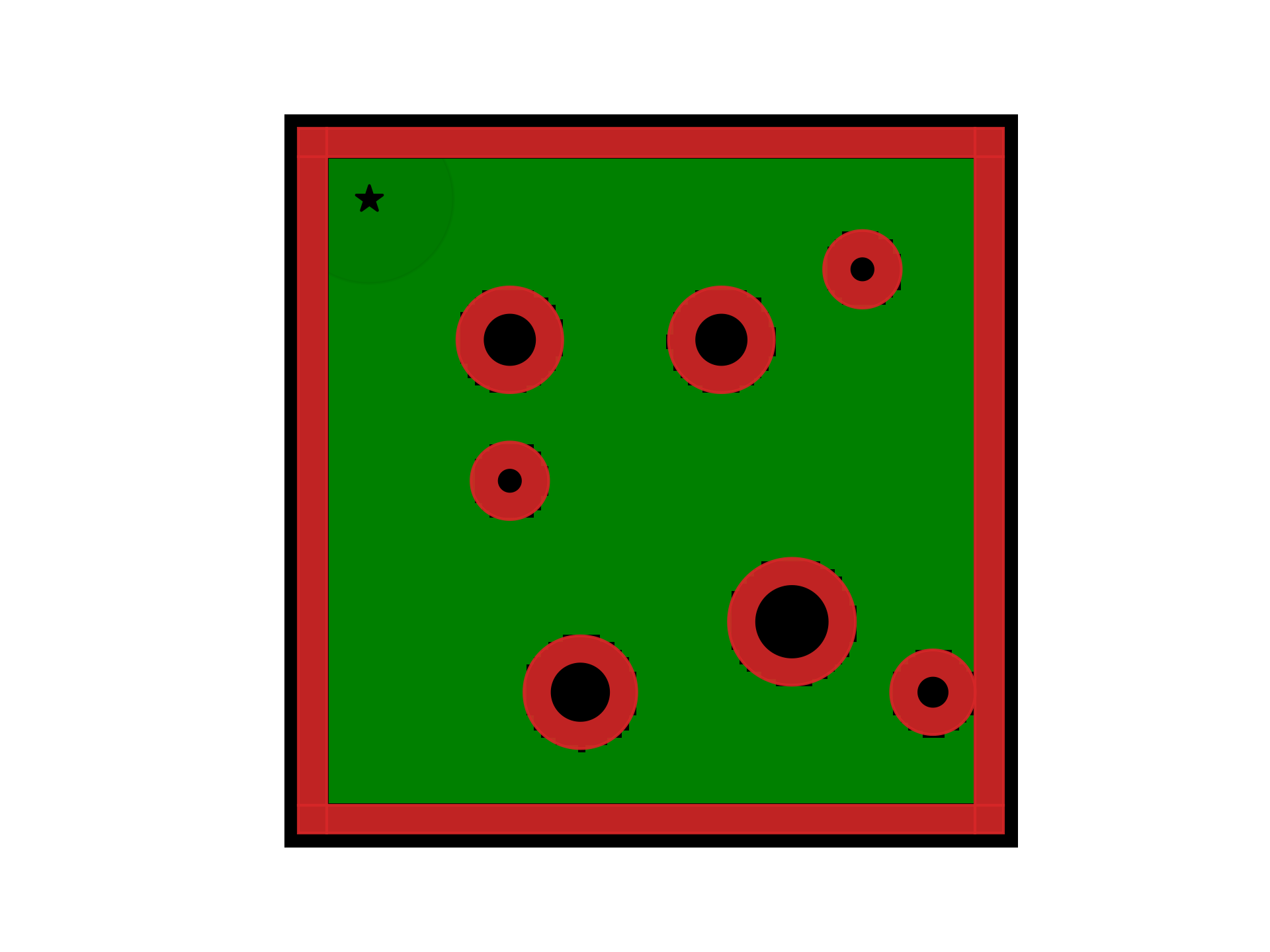} \\
    \includegraphics[trim={3cm 1cm 3cm 0.5cm},clip, width=0.49\linewidth, angle=180,origin=c]{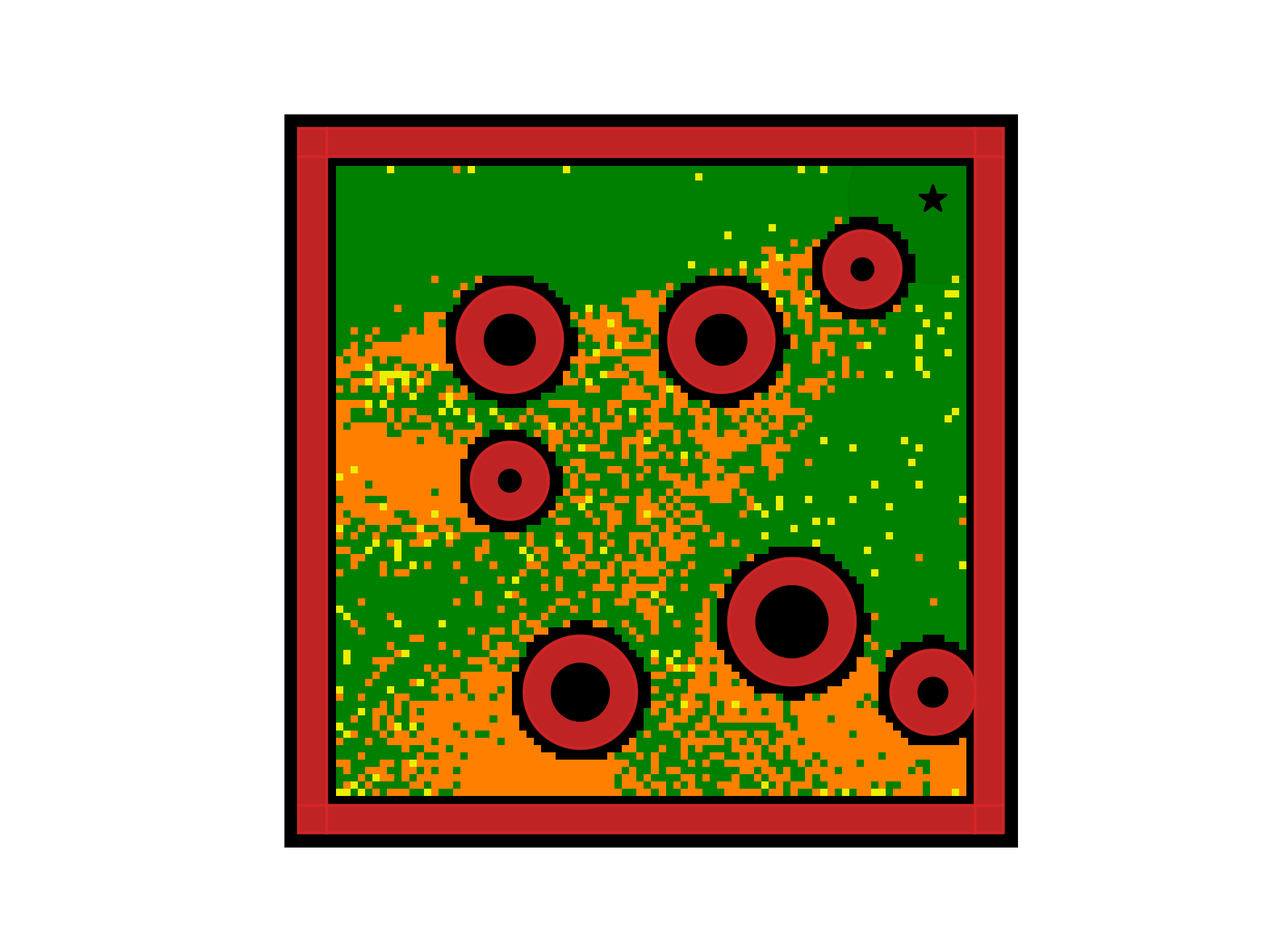}
    \includegraphics[trim={3cm 1cm 3cm 0.5cm},clip, width=0.49\linewidth, angle=180,origin=c]{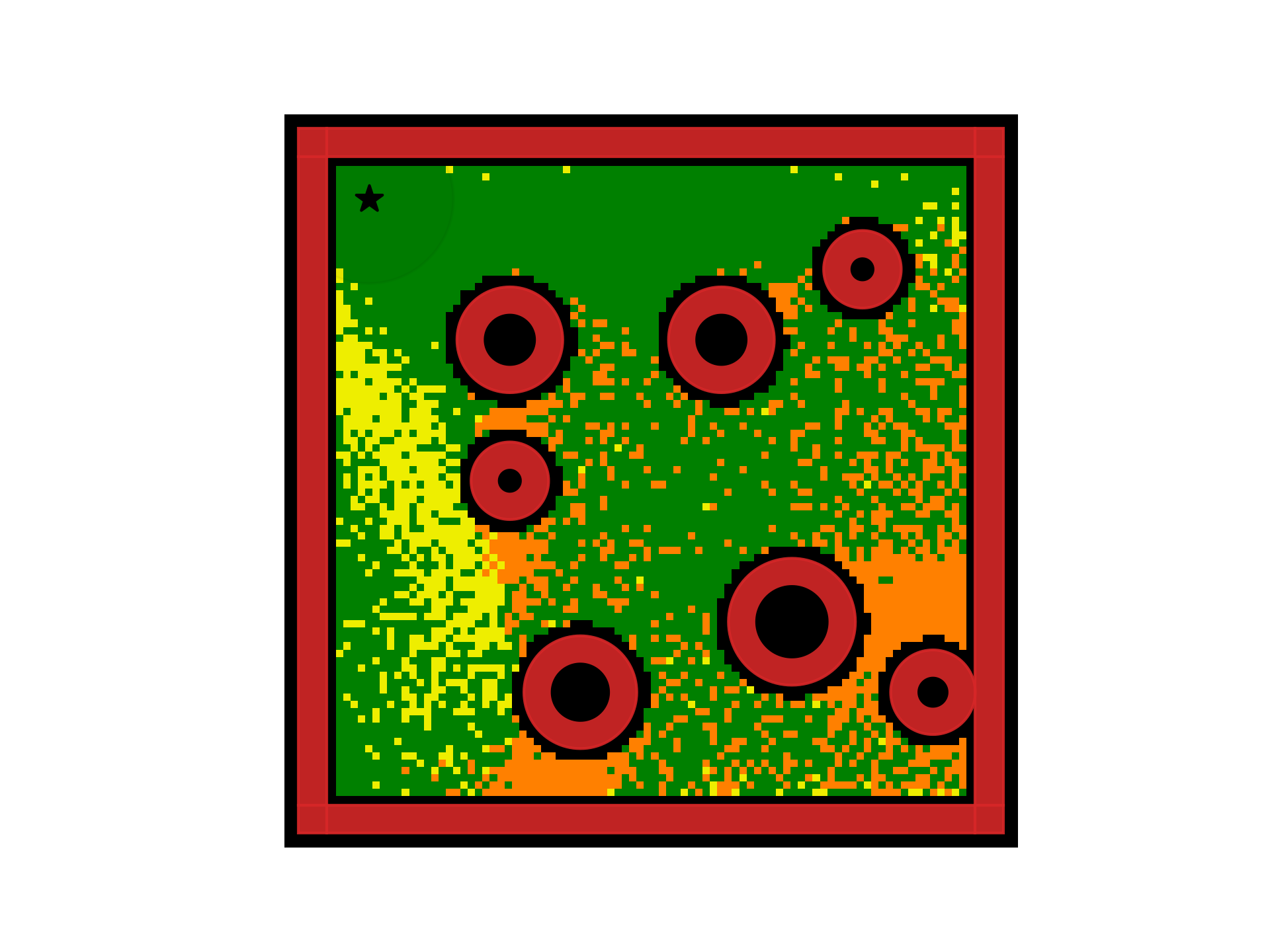} \\
    \includegraphics[trim={3cm 1cm 3cm 0.5cm},clip, width=0.49\linewidth, angle=180,origin=c]{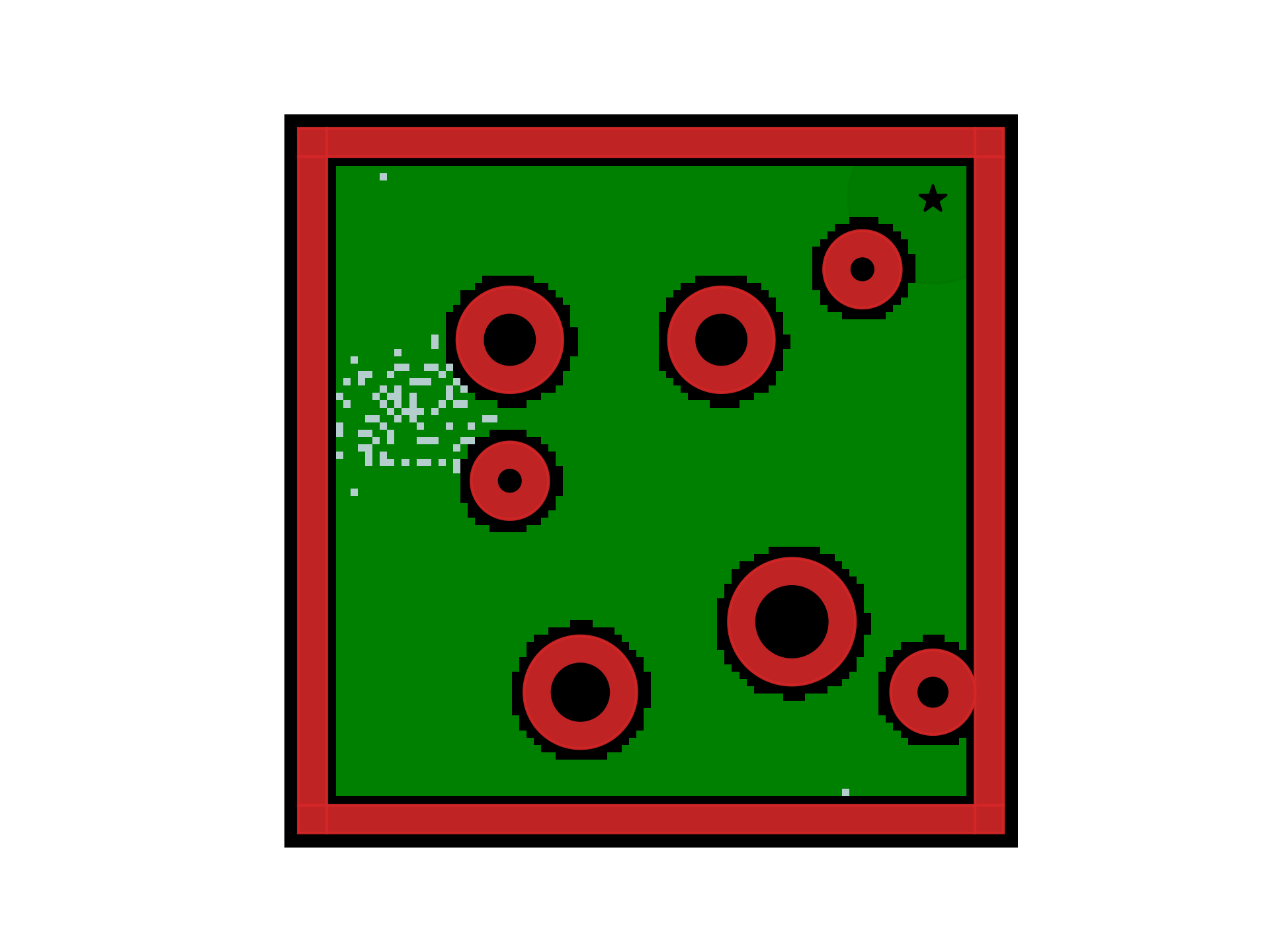}
    \includegraphics[trim={3cm 1cm 3cm 0.5cm},clip, width=0.49\linewidth, angle=180,origin=c]{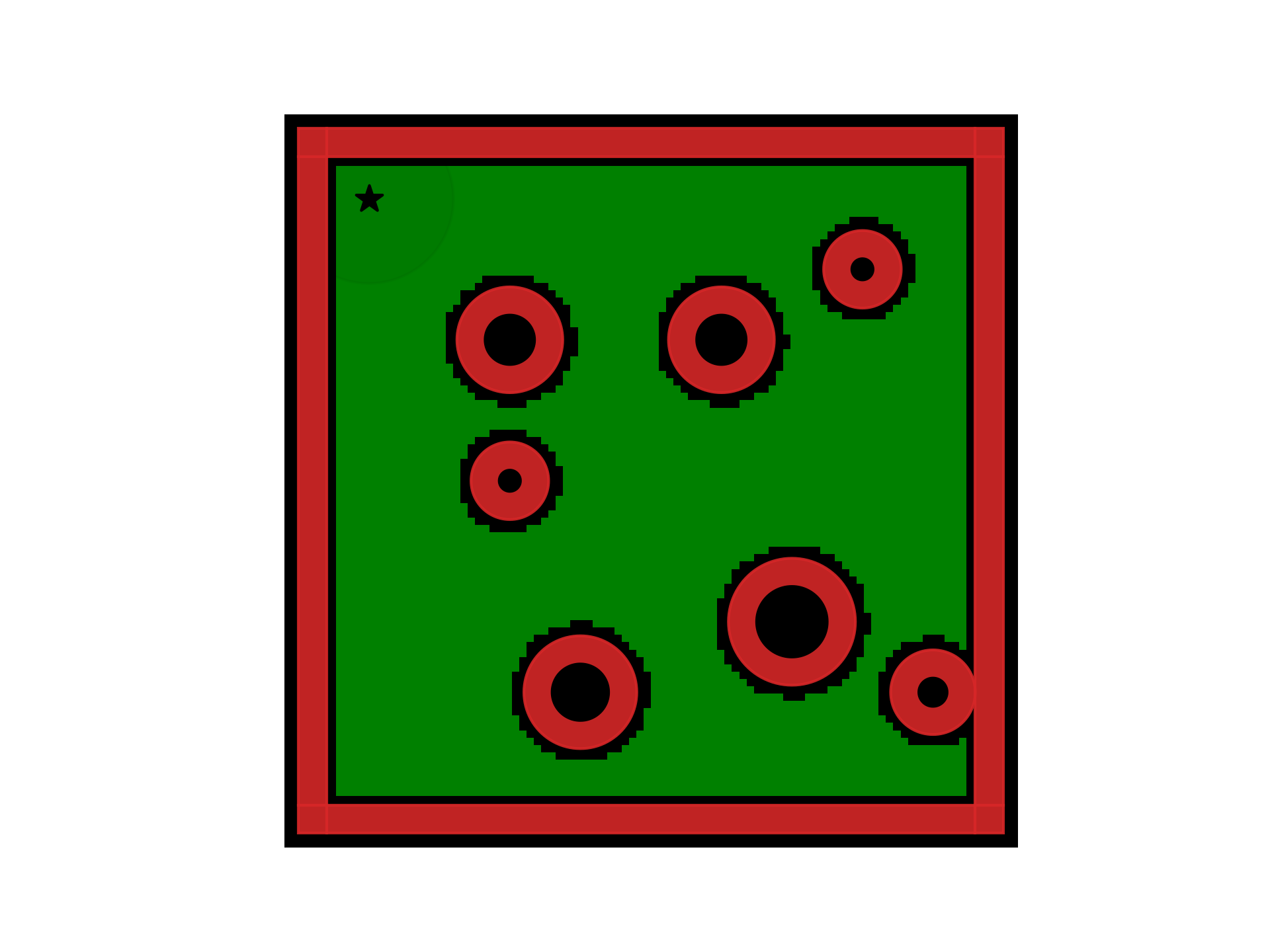}
    \caption{Evaluation of the learned policy for a single agent over a $100\times 100$ grid of initial positions. (Left column) policy for target 1. (Right column) policy for target 2. From top to bottom:
    (Row 1) RL policy, no noise;
    (Row 2) RL+Filter, no noise;
    (Row 3) RL policy, with noise; 
    (Row 4) RL+Filter, with noise. 
    We observe that the combination of safety filter and single-agent RL controller achieves the highest generalization, with and without noise.
    }\label{fig:rl_eval}
\end{figure}

\begin{table*}
    \caption{Comparison of the proposed safety filter \eqref{prob:safety_filter_reform} with a pure MPC-based motion planner \eqref{eq:pure_mpc}. The proposed approach completes the motion planning task for more percentage of trials. 
    We report the $(5,50,95)$-percentiles of the results of the subset of $100$ Monte-Carlo simulations that completed the task successfully.}\label{tab:summary_num}
    \centering
    \begin{tabular}{|c|c||c|c|c|c|}
    \hline
        Safe controller & Term. const. & \% Success & Task
        completion time & Min. obstacle separation & Min. agent separation \\ 
        \hline\hline
        Proposed approach & Yes & 
        99 & (206, 236, 401) &  (0.15, 0.21, 0.26) & (0.31, 0.32, 0.36) \\\cline{2-6}
        RL + Safety filter \eqref{prob:safety_filter_reform} & No &
        92 & (229, 325, 648)  & (0.15, 0.19, 0.24) & (0.25, 0.27, 0.28)  \\\hline\hline
        \multirow{2}{*}{Pure MPC \eqref{eq:pure_mpc}} & Yes & 
        \multicolumn{4}{c|}{Failed at control time step $(10, 10, 10)$}  \\\cline{2-6}
        & No & 
        55 & (110, 150, 194) & (0.15, 0.16, 0.17) & (0.23, 0.23, 0.23) \\\hline
    \end{tabular}
\end{table*}

\subsection{Simulation study: Impact of RL and terminal constraints}

Next, we compare our approach with a pure MPC-based motion
planner in simulation. Specifically, we solved
\eqref{prob:safety_filter_reform}, where the objective
\eqref{eq:optimization_centralized_cost} is replaced
with a set point regulation cost, which results in the optimization problem,
\begingroup
    \makeatletter\def\f@size{9.5}\check@mathfonts
\begin{align}
    \hspace*{-1em}\begin{array}{rl}
    \underset{\{\Uisafe(t)\}_{i=1}^N}{\mathrm{min}}
                &\ \sum_{k=t}^{t+T}\sum_{i=1}^N
                \lambda_{i,k}{\|\mean{p}_i(k|t) - q_i \|}^2
                + \varepsilon \|\uisafe(k|t)\|^2,\\
            \mathrm{s.\ t.} 
                &\ \text{Constraints of \eqref{prob:safety_filter_reform}},
    \end{array}\label{eq:pure_mpc}
\end{align}
\endgroup
with $\Uisafe(t)=\alluisafe$ for each $i\in\Nint{1}{N}$, pre-specified weights $\lambda_{i,t}\geq 0$ on the
deviations ${\|\mean{p}_i(k|t) - q_i \|}^2$, and a penalty for inputs $\varepsilon>0$.

Problem \eqref{eq:pure_mpc} is a convex quadratic program, thanks to the convexification step
(Propositions~\ref{prop:risk_tightened_constraints} and~\ref{prop:risk_tightened_terminal_constraints}) that uses a
modified version of \eqref{eq:convexification_vec}. Recall that the constraints of \eqref{prob:safety_filter_reform}
included constraints \eqref{eq:prop_1_eqns} and \eqref{eq:prop_2_eqns} that required user-specified unit vectors
$\zijobs,\zijagent,\ellijobs$, and $\ellijagent$, which were defined using the RL trajectory in
\eqref{eq:convexification_vec}.  When formulating \eqref{eq:pure_mpc}, we defined these vectors using the baseline
controller trajectory instead of the RL trajectory for a fair comparison.
One can view \eqref{eq:pure_mpc} as an extension of existing single-agent motion planners under uncertainty (for
example,~\cite{blackmore2011chance,vinod2018stochastic}) for multi-agent motion
planning, with the addition of terminal constraints for
recursive feasibility proposed in Section~\ref{sub:recursive_feasibility_reform}. Note that \eqref{eq:pure_mpc} enables
explicit coordination between agents as they move towards
their goal, while the proposed safety filter
\eqref{prob:safety_filter_reform} only minimizes deviations
from RL-based single-agent motion planners. We now study
the RL block and the terminal constraints
(Proposition~\ref{prop:risk_tightened_terminal_constraints})
by comparing the performances of the proposed approach and a
pure MPC approach \eqref{eq:pure_mpc}, with and without
terminal constraints
(Proposition~\ref{prop:risk_tightened_terminal_constraints}). 

Table~\ref{tab:summary_num} summarizes the performance of
both the approaches in $100$ Monte-Carlo simulations. We
observe that the proposed approach
\eqref{prob:safety_filter_reform} completed the motion
planning task for a significantly larger number of
simulations than a pure MPC approach \eqref{eq:pure_mpc}
($99\%$ vs $55\%$ success), illustrating the benefits of
including RL. The sources of failure in these simulations include collisions with static or dynamic obstacles (safety is
enforced in probability) as well as numerical issues for the solver.
For the proposed approach \eqref{prob:safety_filter_reform},
the use of terminal constraints for recursive feasibility
(Proposition~\ref{prop:risk_tightened_terminal_constraints})
typically resulted in a larger minimum separation between
agents and obstacles, and among agents. The use of terminal
constraints also led to smaller task completion time,
possibly due to the larger minimum separations. On the other
hand, the use of similar terminal constraints in the MPC approach
\eqref{eq:pure_mpc} made the problem considerably harder and
led to numerical issues in all trials, possibly because the trajectory of the
baseline controller may not be as informative as the RL trajectory for the convexification step. Finally, we observe that the
proposed approach takes longer to complete the motion
planning task than the pure MPC approach without the terminal constraints, when the latter
does not result in safety violations. This is expected since the terminal constraints impose additional
restriction on the generated trajectory to achieve recursive feasibility. The single agent
motion planner combined with safety filter is suboptimal
when applied to a multi-agent motion planning problem and is more conservative due to the terminal
constraints, but
guarantees safety. Thus, there is a trade-off between safety
and performance.

\subsection{Scalability study of the proposed approach}
\begin{figure}[t]
    \centering
    \includegraphics[width=1\linewidth]{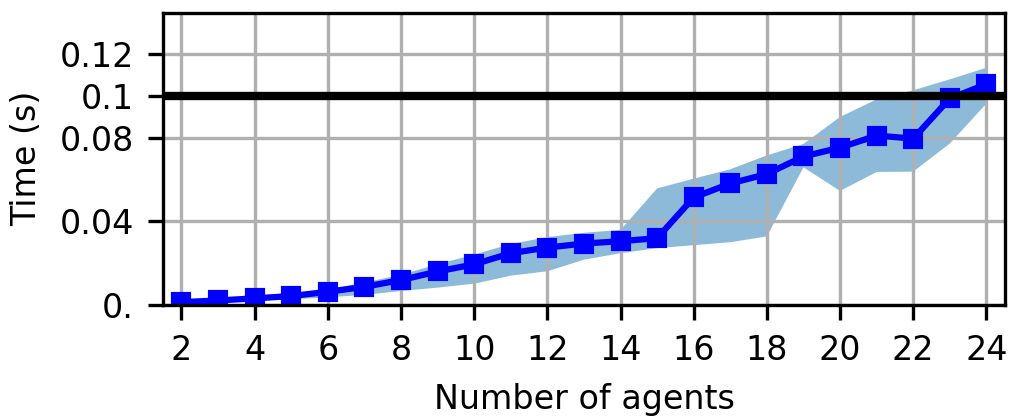} 
    \vspace*{-1em}
    \caption{Computation times (in $(5,50,95)$ percentiles) of the safety filter show a modest increase as the number of
agents increases. The computation times were collected from $1000$ control time steps of the simulated workspace. We used \texttt{GUROBI}~\cite{GUROBI} to solve the quadratic program \eqref{prob:safety_filter_reform}.
}\label{fig:scalability}
\end{figure}

To perform scalability analysis of the safety filter, we
reduced $r_A$ to $0.01$, reduced the noise covariance from $10^{-4}$ to $10^{-6}$, and collected computational times
for the safety filter for $1000$ control time steps starting from
randomly initialized locations for the agents in simulation. 

Figure~\ref{fig:scalability} shows the computation time to
solve  \eqref{prob:safety_filter_reform}, where the number of
agents ranges from $2$ to $24$. The compute time of the
safety filter increases only moderately with the number of
agents, thanks to the convex quadratic program structure of
\eqref{prob:safety_filter_reform}. Compared to our preliminary
work in the deterministic setting~\cite{our_icra},
\eqref{prob:safety_filter_reform} needs a larger computational
effort, possibly due to the larger number of decision
variables and larger number of constraints. Specifically,
\eqref{prob:safety_filter_reform} computes time-varying
control commands over the planning horizon
compared to constant input approach used in~\cite{our_icra},
and \eqref{prob:safety_filter_reform} includes additional
constraints for recursive feasibility \eqref{eq:prop_2_eqns}. 

\section{Conclusion}
We presented a solution for the multi-agent motion planning problem that combines reinforcement learning and
constrained control. We utilize single-agent RL to train a policy for traversing a cluttered workspace while ignoring inter-agent collision avoidance, and use a real-time implementable, constrained-control-based safety filter to account for inter-agent collision avoidance and ensure probabilistic collective safety of the agents. The formulated QP includes chance constraints to achieve safety under process and measurement noise as well as probabilistic recursive feasibility constraints. We demonstrated the efficacy of our approach via numerical simulations, and validated our approach on a hardware testbed using quadrotors. 

{In our future work, we will investigate the application of the proposed approach in a decentralized setting, consider safe multi-agent motion planning for agents with nonlinear dynamics, and evaluate RL-based planning with a subset of the multiple agents larger than one.}

\appendix
{\hfill\textsc{Proof of Proposition~\ref{prop:risk_tightened_constraints}}\hfill}\label{app:risk_tightened_constraints_proof}

\textit{Static obstacle collision avoidance} (\eqref{eq:prop_1_eqns_obs} $\Rightarrow$
\eqref{eq:prob_collect_static_obstacle_cc})\emph{:}
    Using computational geometry arguments, \eqref{eq:prob_collect_static_obstacle_cc} is a non-convex chance
    constraint, and is equivalent to
    \begin{align}
        \bbp(\rv{p}_i(k|t) - \rv{c}_j(t) \not\in \obst{j} \oplus (-\agent)) \geq 1-\alpha_{i,j,t}.\label{eq:static_obst_reform_0}
    \end{align}
    To convexify it, we use a separating hyperplane for $(\mean{p}_i(k|t) - \mean{c}_j)$ and $\obst{j} \oplus (-\agent)$
    along the direction of a user-specified direction $\zijobs$~\cite{boyd2004convex}.
    Thus, 
    \begin{align*}
        &\bbp(\zijobs \cdot ({\rv{p}}_i(k|t) - {\rv{c}}_j) \geq S_{\obst{j}}(\zijobs) + S_{-\agent}(\zijobs)) \geq 1-\alpha_{i,j,t}  \\
        &\Longleftrightarrow \bbp(\zijobs \cdot ({\rv{p}}_i(k|t) - {\rv{c}}_j) \leq S_{\obst{j}}(\zijobs) + S_{-\agent}(\zijobs)) \leq \alpha_{i,j,t}  \\
        &\ \Longrightarrow \eqref{eq:static_obst_reform_0}.
    \end{align*}
    We use \eqref{eq:simple_cc_reform1} in Lemma~\ref{lemma:gauss}  to reformulate the left hand side of the above implication to arrive at
    \eqref{eq:prop_1_eqns_obs}. Thus, \eqref{eq:prob_collect_static_obstacle_cc} holds, if
    \eqref{eq:prop_1_eqns_obs} holds.

    \textit{Inter-agent collision avoidance} (\eqref{eq:prop_1_eqns_agent} $\Rightarrow$
    \eqref{eq:prob_collect_inter_agent_cc})\emph{:} Using arguments similar to the
    above with $\zijagent, \mean{p}_j(k|t), \Sigma_{p_j}(k|t)$ instead of $\zijobs, \mean{c}_j, \Sigma_{c_j}$, we can
    show that \eqref{eq:prob_collect_inter_agent_cc} holds, if \eqref{eq:prop_1_eqns_agent} holds.

    \textit{Keep-in constraint} (\eqref{eq:prop_1_eqns_env} $\Rightarrow$
    \eqref{eq:prob_collect_keepin_cc})\emph{:}
    From the definition of {Pontryagin} difference, \eqref{eq:prob_collect_keepin_cc} is equivalent to
    \begin{align}
         \bbp(\rv{p}_i(k|t) \not\in \envPoly \ominus \agent) \leq \kappa_{i,t} \label{eq:keep_in_reform_1}.
    \end{align}
    Here, $\envPoly \ominus \agent$ is easy to compute~\cite[Thm~2.3]{kolmanovsky1998theory}. Specifically, 
        $\envPoly \ominus \agent = \cap_{i \in \Nint{1}{\numEnvHalfspaces}} \{p : \envHalfspaceA_i \cdot p \leq \envHalfspaceB_i - S_{\agent}(\envHalfspaceA_i)\}$.
    Using Boole's inequality and assuming that the risk bound is divided equally across all halfspaces, we have
    \begin{align*}
        &\bbp(\envHalfspaceA_j \cdot \rv{p}_i(k|t) > \envHalfspaceB_j - S_{\agent}(\envHalfspaceA_j)) \leq \frac{\kappa_{i,t}}{\numEnvHalfspaces}, \ \forall j \in \Nint{1}{\numEnvHalfspaces} \Rightarrow \eqref{eq:keep_in_reform_1}.
    \end{align*}
    We use \eqref{eq:simple_cc_reform2} in Lemma~\ref{lemma:gauss}  to reformulate the left hand side of the above
    implication to arrive at \eqref{eq:prop_1_eqns_env}. Thus, \eqref{eq:prob_collect_keepin_cc} holds, if
    \eqref{eq:prop_1_eqns_env} holds.\hfill\qed

\bibliography{refs}
   
\end{document}

%% file: commands.tex
\newcommand{\bbe}{\mathbb{E}}
\newcommand{\bbn}{\mathbb{N}}
\newcommand{\bbp}{\mathbb{P}}
\newcommand{\bbr}{\mathbb{R}}

\newcommand{\cala}{\mathcal{A}}

\newcommand{\calc}{\mathcal{C}}

\newcommand{\calf}{\mathcal{F}}
\newcommand{\calk}{\mathcal{K}}
\newcommand{\caln}{\mathcal{N}}
\newcommand{\calo}{\mathcal{O}}

\newcommand{\calu}{\mathcal{U}}

\newcommand{\scra}{\mathscr{A}}

\newcommand{\scrv}{\mathscr{V}}

\newcommand{\Nint}[2]{\mathbb{N}_{[#1, #2]}}

\newcommand{\usafe}{u^\mathrm{safe}}
\newcommand{\Usafe}{U^\mathrm{s}}
\newcommand{\uisafe}{\usafe_i}
\newcommand{\Uisafe}{\Usafe_i}
\newcommand{\alluisafe}{\{{\usafe_i(k|t)}\}_{k=t}^{t+T-1}}
 
\newcommand{\uirl}{u_i^\mathrm{RL}}
\newcommand{\zijobs}{z_{ij}^\text{obs}}
\newcommand{\zijagent}{z_{ij}^\text{agt}}
\newcommand{\ellijobs}{\ell_{ij}^\text{obs}}
\newcommand{\ellijagent}{\ell_{ij}^\text{agt}}

\usepackage{bm}
\newcommand*\rv[1]{\bm{#1}}  
\newcommand*\mean[1]{\overline{#1}}  
\newcommand*\estrealize[1]{\hat{#1}}  

\newcommand{\Phiinv}{\Phi^{-1}}

\newcommand{\inputSet}{\calu}

\newcommand{\horizon}{T}

\newcommand*\cvxCompactSet{\calc}  
\newcommand*\envPoly{\calk}  
\newcommand*\numEnvHalfspaces{N_\calk}  
\newcommand*\envHalfspaceA{h}  
\newcommand*\envHalfspaceB{g}  
\newcommand*\agentRad{r}  
\newcommand*\agent{\cala}  
\newcommand*\numAgents{N}  
\newcommand*\obst[1]{\calo_{#1}}  
\newcommand*\numObst{N_O}  
	
\newcommand{\gauss}[2]{\caln(#1 #2)}

\newtheorem{rem}{Remark}
\newtheorem{lemma}{Lemma}

\newtheorem{prob}{Problem}
\newtheorem{prop}{Proposition}
\newtheorem{defn}{Definition}

\usepackage{xcolor}

\newcommand{\AvoidSet}{\operatorname{AvoidSet}}
\newcommand{\AvoidSetP}{\operatorname{AvoidSet}^+}
\newcommand{\ViabSet}{\operatorname{ViabilitySet}}

\newcommand{\ellipAvoid}[1]{\scra_{#1}}
\newcommand{\polyViab}{\scrv}
\newcommand*\numPolyViabHalfspaces{N_\polyViab}  
\newcommand*\polyViabHalfspaceA{\underline{h}}  
\newcommand*\polyViabHalfspaceB{\underline{g}}  

\newcommand{\nnet}{\nu}
\newcommand{\rlmotionplan}{\{x^{\text{RL}}_i(k|t)\}_{k=t}^{t+T}}
\newcommand{\rlcontrolseq}{\{u^{\text{RL}}_i(k|t)\}_{k=t}^{t+T-1}}